\documentclass[letterpaper, 10 pt, journal]{IEEEtran}
\usepackage{amsmath,amsfonts}
\usepackage{textcomp}
\usepackage{stfloats}
\usepackage{verbatim}
\usepackage{graphicx}
\usepackage{url}
\usepackage{pifont}
\usepackage{cases}
\usepackage{multirow}
\usepackage{ulem}
\usepackage{bbm}
\usepackage{arydshln}
\def\BibTeX{{\rm B\kern-.05em{\sc i\kern-.025em b}\kern-.08em
    T\kern-.1667em\lower.7ex\hbox{E}\kern-.125emX}}
\usepackage{balance}
\usepackage{cite}
\usepackage{booktabs}
\begin{document}
\title{Hyperbolic Distillation: Geometry-Guided Cross-Modal Transfer for Robust 3D Object Detection}
\author{Kanglin Ning, Wenrui Li, Houde Quan, Qifan Li, Xingtao Wang, Xiaopeng Fan , \textit{Senior Member, IEEE}
\thanks{This work was supported in part by the National Key R\&D Program of China (2025ZD1601300), the National Natural Science Foundation of China (NSFC) under grants 62402138 and 624B2049, and the Fundamental and Interdisciplinary Disciplines Breakthrough Plan of the Ministry of Education of China (JYB2025XDXM901), and the Suzhou Key Core Technology Project under grant number SYG2025118. (Corresponding author: Xiaopeng Fan.)}
\thanks{Kanglin Ning, Houde Quan, Qifan Li, Wenrui Li, Xingtao Wang are with the Faculty of Computing, Harbin Institute of Technology, Harbin 150001, China. Kanglin Ning and Xingtao Wang are also currently affiliated with the Suzhou Research Institute of HIT. (email: 23B936010@stu.hit.edu.cn; liwr618@163.com;23b903005@stu.hit.edu.cn; 23B903056@stu.hit.edu.cn;  xtwang@hit.edu.cn)}
\thanks{Xiaopeng Fan is with the Faculty of Computing, Harbin Institute of Technology, Harbin 150001, China. He is also with the PengChengLab, Shenzhen 518055, China, and the Suzhou Research Institute of HIT. (e-mail: fxp@hit.edu.cn).}
}

\markboth{IEEE Transactions on Multimedia}%
{Hyperbolic-Constraint 2D Prior Guided Multi-Modal 3D object Detection}

\maketitle

\begin{abstract}
Cross-modal knowledge distillation has emerged as an effective strategy for integrating point cloud and image features in 3D perception tasks. However, the modality heterogeneity, spatial misalignment, and the representation crisis of multiple modalities often limit the efficient of these cross-modal distillation methods. To address these limitations in existing approaches, we propose a hyperbolic constrained cross-modal distillation method for multimodal 3D object detection (HGC-Det). The proposed HGC-Det framework includes an image branch and a point cloud branch to extract semantic features from two different modalities. The point cloud branch comprises three core components: a 2D semantic-guided voxel optimization component (SGVO), a hyperbolic geometry constrained cross-modal feature transfer component (HFT), and a feature aggregation-based geometry optimization component (FAGO). Specifically, the SGVO component adaptively refines the spatial representation of the 3D branch by leveraging semantic cues from the image branch, thereby mitigating the issue of inadequate representation fusion. The HFT component exploits the intrinsic geometric properties of hyperbolic space to alleviate semantic loss during the fusion of high-dimensional image features and low-dimensional point cloud features. Finally, the FAGO compensates for potential spatial feature degradation introduced by the 2D semantic-guided voxel optimization component. Extensive experiments on indoor datasets (SUN RGB-D, ARKitScenes) and outdoor datasets (KITTI, nuScenes) demonstrate that our method achieves a better trade-off between detection accuracy and computational cost.

\end{abstract}

\begin{IEEEkeywords}
3D Object Detection, Multi-Modal Learning, Cross-Modal Knowledge Distillation
\end{IEEEkeywords}

\section{Introduction}
As single image or LiDAR point cloud cannot provide a panoramic or dense environmental scan \cite{xu2025survey}, multi-sensor fusion is essential for achieving robust and precise 3D perception performance.

Existing 3D object detection methods that fuse image and point cloud data are typically based on lift-splat-shot strategies\cite{yin2024fusion, liang2022bevfusion, li2022voxel} or cross-attention modules\cite{wang2025mv2dfusion, li2023logonet}.  While these methods have yielded substantial gains in fusion accuracy, their considerable computational overhead inevitably elevates the hardware cost required for practical deployment. To mitigate this limitation, the Fusion-to-single cross-modal distillation paradigm\cite{yan20222dpass, yuan2022x, 10659158} has emerged as a compelling alternative. These methods typically fuse image and point cloud features and then distill the fused representations into point cloud features, a process that inherently requires channel-level compression of high-dimensional image features and inevitably results in semantic information loss (as shown in Fig. 1). Moreover, owing to the inherent sparsity of point cloud data, existing cross-modal distillation methods often exhibit suboptimal utilization of image features (as shown in Fig. 2).

\begin{figure}
\centering
\includegraphics[width=3.3in]{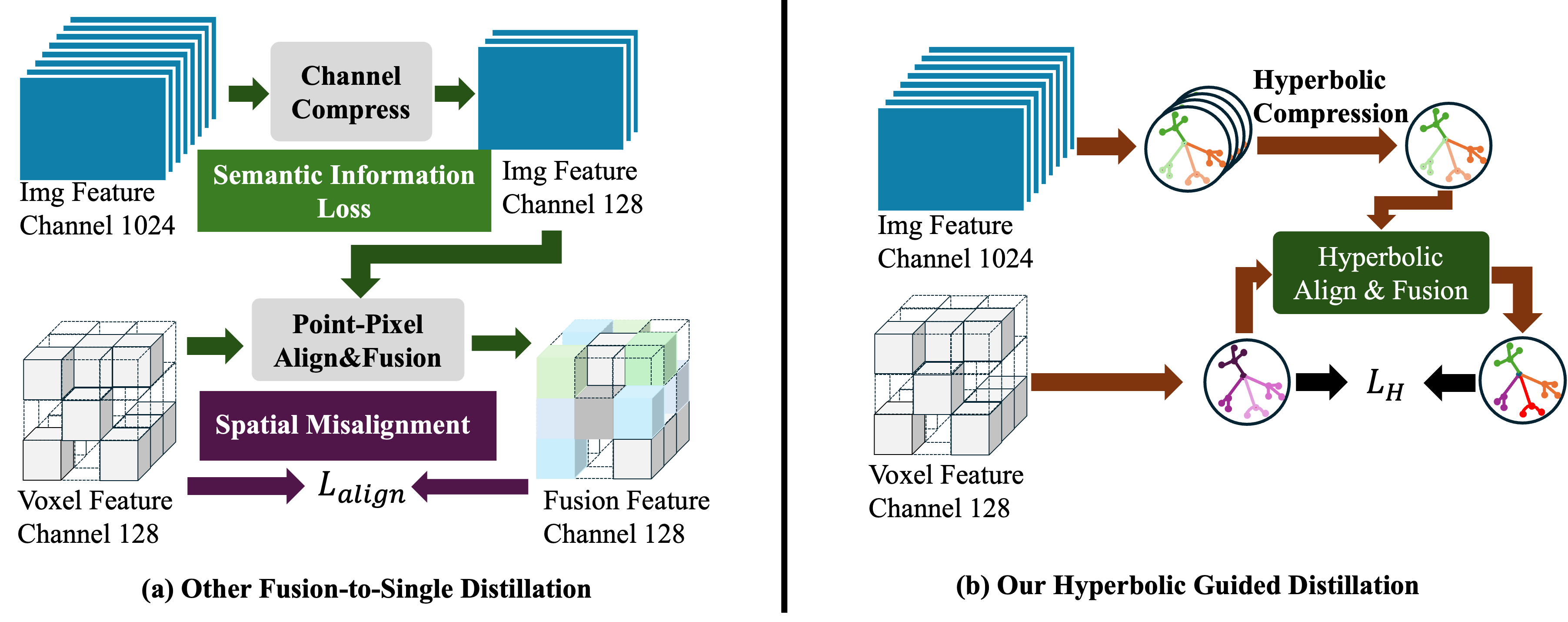}
\caption{A comparative illustration of the conventional Fusion-to-Single cross-modal distillation pipeline and our proposed Hyperbolic Guided distillation framework is provided.}
\setlength{\abovecaptionskip}{-0.3cm}
\end{figure}

The issue of semantic information loss can be further elucidated through an intuitive illustrative example. Prior research \cite{dong2025global, luo2024revisiting, xu2025parameterized, li2024superpixel} indicates that deep networks capture spatial patterns in shallow, low-dimensional layers, while semantic information is encoded in deep, high-dimensional layers. Consequently, categories with similar geometry are indistinguishable in low-dimensional spaces but become separable in high-dimensional ones. A straightforward compression module applied to high-channel features in Euclidean space inevitably degrades this separability and discards the original semantic hierarchy. For brevity, we refer to this phenomenon as the loss of semantic structural hierarchy under Euclidean compression.

The issue of insufficient image feature utilization is illustrated in Fig. 2, which depicts the projection of both the original and voxelized point clouds onto the corresponding 2D image plane. In outdoor scenarios, LiDAR-acquired point clouds are inherently sparse, such that their projection from 3D space onto the 2D image plane occupies only a minimal fraction of the available pixel area. Consequently, existing Fusion-to-single cross-modal distillation methods inevitably incur a spatial misalignment between point clouds and image pixels.

\begin{figure}
\centering
\includegraphics[width=3.0in]{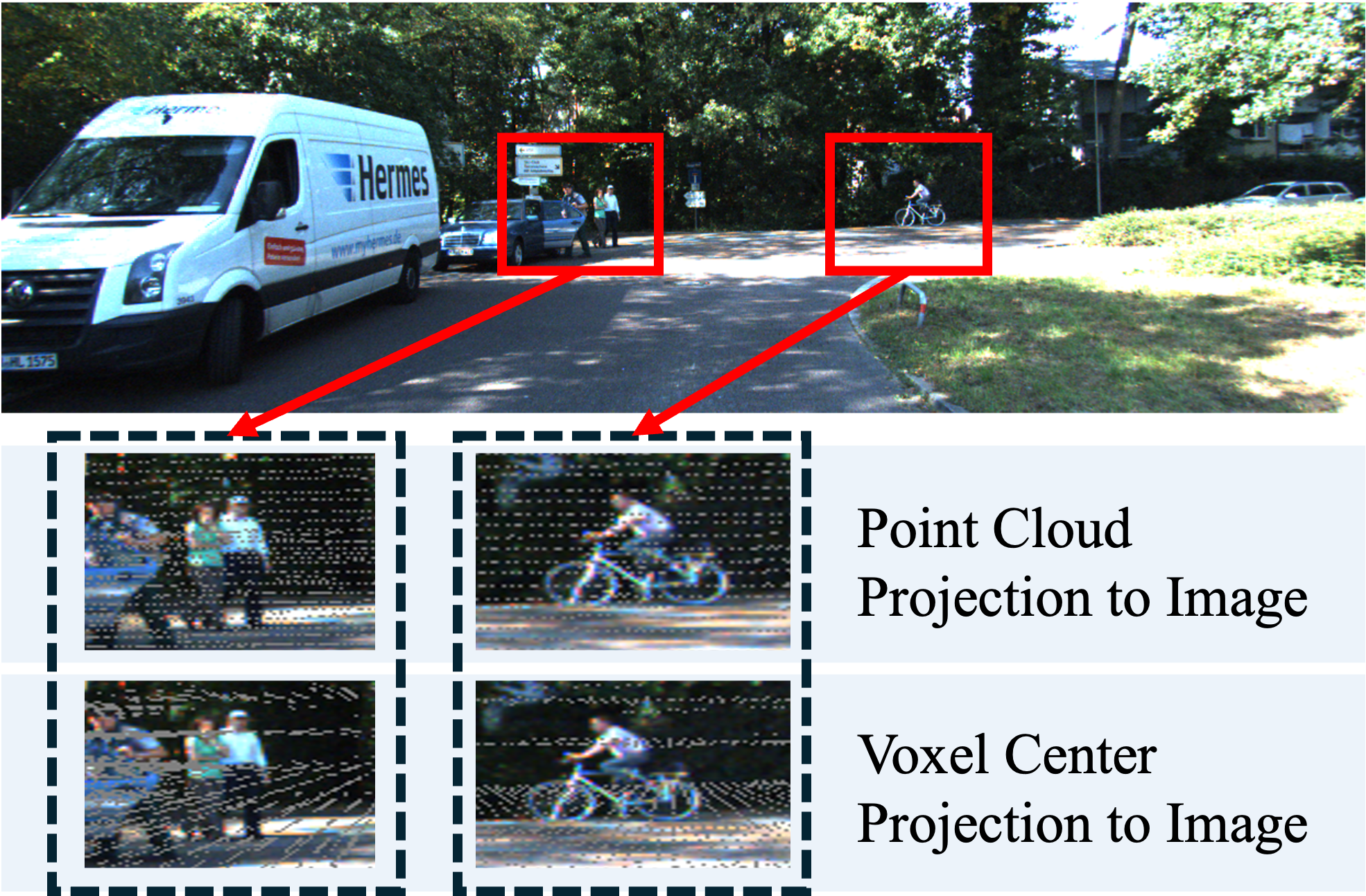}
\caption{An illustrative example of projecting a 3D point cloud onto a 2D image plane is presented. The first row displays the corresponding input image. The second row shows the projection of the original point cloud onto the image plane, while the third row depicts the projection of the center points of non-empty voxels—obtained after voxelization—onto the same image plane.}
\end{figure}

To address two fundamental limitations of existing approaches\cite{yan20222dpass, yuan2022x, 10659158}, we propose a hyperbolic constrained cross-modal distillation method for multimodal 3D object detection, termed HGC-Det. The proposed HGC-Det framework includes an image branch and a point cloud branch to extract semantic features from two different modalities. The point cloud branch comprises three core components: a 2D semantic-guided voxel optimization component (SGVO), a hyperbolic geometry constrained cross-modal feature transfer component (HFT), and a feature aggregation-based geometry optimization component (FAGO). The SGVO module is primarily designed to mitigate the spatial misalignment of image-point features. This module leverages semantic information from the image branch to partition the 3D feature space into foreground and background regions, and adaptively refines their spatial distributions to mitigate spatial misalignment while maintaining overall efficiency. The HFT component performs compression, fusion, and cross-modal distillation of high-channel-dimensional image features directly in hyperbolic space, exploiting its inherent capacity to preserve hierarchical structure and amplify inter-sample distances, thereby mitigating the loss of semantic structural hierarchy during feature compression. Finally, the FAGO component serves to compensate for geometric structural distortion potentially introduced during the voxel space redistribution performed by SGVO, ensuring that the overall geometric fidelity of the 3D representation is maintained. In general, our contributions can be summarized into the following three points:

\begin{itemize}
\item We propose a Hyperbolic Guided Cross-modal Multi-modal 3D object detection framework to tackle the issue of semantic structural hierarchy loss and spatial misalignment problems.
\item We design the HFT component to leverage the intrinsic advantages of hyperbolic space, thereby substantially mitigating the loss of semantic structural information during the fusion of image and point cloud features.
\item We design the SGVO component to tackle the spatial misalignment between point cloud and image by exploiting 2D semantic information from the image branch to adaptively refine the spatial distribution of 3D representations.
\item We design the FAGO component to recover geometric structure lost by SGVO by reconstructing the geometric center of each region from a dense 3D representation through center voting over both feature and geometric dimensions.
\end{itemize}

\section{Related Works}

\subsection{Multi-Modal 3D object Detection}
The goal of multimodal 3D object detection methods is to fuse visual input data from different sensors to achieve comprehensive and high-density perception and understanding of the surrounding environment\cite{pravallika2024deep, wu2025fully, hou2025binarized, wu2024inf}. Existing multimodal fusion methods\cite{rukhovich2023tr3d, wang2025mv2dfusion, yin2024fusion, yan2023cross,  liang2022bevfusion, wang2023unitr, song2023simple, li2023logonet, 10510171} typically consider fusing point cloud and image data. These methods are categorized into indoor\cite{rukhovich2023tr3d, qi2020imvotenet} and outdoor scenarios\cite{ wang2025mv2dfusion, yin2024fusion, yan2023cross, liang2022bevfusion, wang2023unitr, li2023logonet, 10510171}. For outdoor scenes, current methods can be broadly categorized into viewpoint transformation-based approaches \cite{yin2024fusion, xie2023sparsefusion, liang2022bevfusion, song2023simple, 10510171}, which rely on coordinate transformation matrices and estimated depth distributions to project image features into BEV space, and attention-based fusion strategies \cite{wang2025mv2dfusion, yan2023cross, li2023logonet, wang2023unitr} that employ cross-attention mechanisms for learnable modality integration.  In contrast, indoor point clouds \cite{song2015sun, baruch2021arkitscenes} exhibit tighter image-point correspondence, and indoor fusion methods \cite{rukhovich2023tr3d, qi2020imvotenet} accordingly prioritize voxel-level aggregation with less concern for viewpoint shift. This yields lower computational cost but renders such approaches less effective in outdoor settings. Different from existing indoor-outdoor scene fusion strategies, our method combines the computationally resource-friendly advantages of indoor scene fusion strategies with good performance in outdoor scenes.

\subsection{Cross-Modal Knowledge Distillation}

The initial goal of knowledge distillation\cite{hinton2015distilling} was to transfer knowledge learned from a large-capacity teacher model to a small-capacity student model, thereby achieving model compression. In recent years, with the development of multi-modal compute vision, some recent methods\cite{yan20222dpass, yuan2022x, liu2021learning, xu2021image2point, 10659158} have attempted to achieve knowledge transfer between different modalities. This involves transferring knowledge extracted from the image modality to the detector in the point cloud modality during the training phase, thereby enhancing performance during inference. \cite{liu2021learning} introduces the 2D-assisted pre-training, \cite{xu2021image2point} inflates the kernels of 2D convolution to the 3D ones, and \cite{yuan2022x} applies a well-designed teacher-student framework. Specifically, \cite{yan20222dpass, 10659158} designed an attention-based fusion-to-single transfer strategy, which demonstrates superior performance compared to methods that rely on pixel matching relationships in point cloud and images. However, while prevailing cross-modal knowledge distillation methods leverage viewpoint transformers for fusing point cloud and image features, they tend to disregard two critical issues: (1) the insufficient fusion arising from inherent modality discrepancies, and (2) the loss of high-dimensional semantic hierarchical structure in images during the fusion stage.

\subsection{Hyperbolic Geometry}

Hyperbolic geometry becomes an effective method for constructing structured representations due to its natural advantage in encoding hierarchical structures \cite{fang2023hyperbolic}. It has wide applications in various fields of computer vision; hyperbolic embedding enhances tasks such as semantic segmentation\cite{atigh2022hyperbolic}, anomaly recognition\cite{hong2023curved} and audio-visual question answering\cite{yang2025shmamba}, demonstrating its potential for capturing the hierarchical structure of visual data. Recent work \cite{sur2025hyperbolic, liu2025hyperbolic, li2025hyperbolic}  has also attempted to leverage the unique advantages of hyperbolic space in 3D point cloud research. In this work, we leverage the geometric properties of hyperbolic space to alleviate the loss of semantic structural hierarchy in existing cross-modal distillation.

\section{Methodology}
\begin{figure*}[!t]
\centering
\includegraphics[width=7.0in]{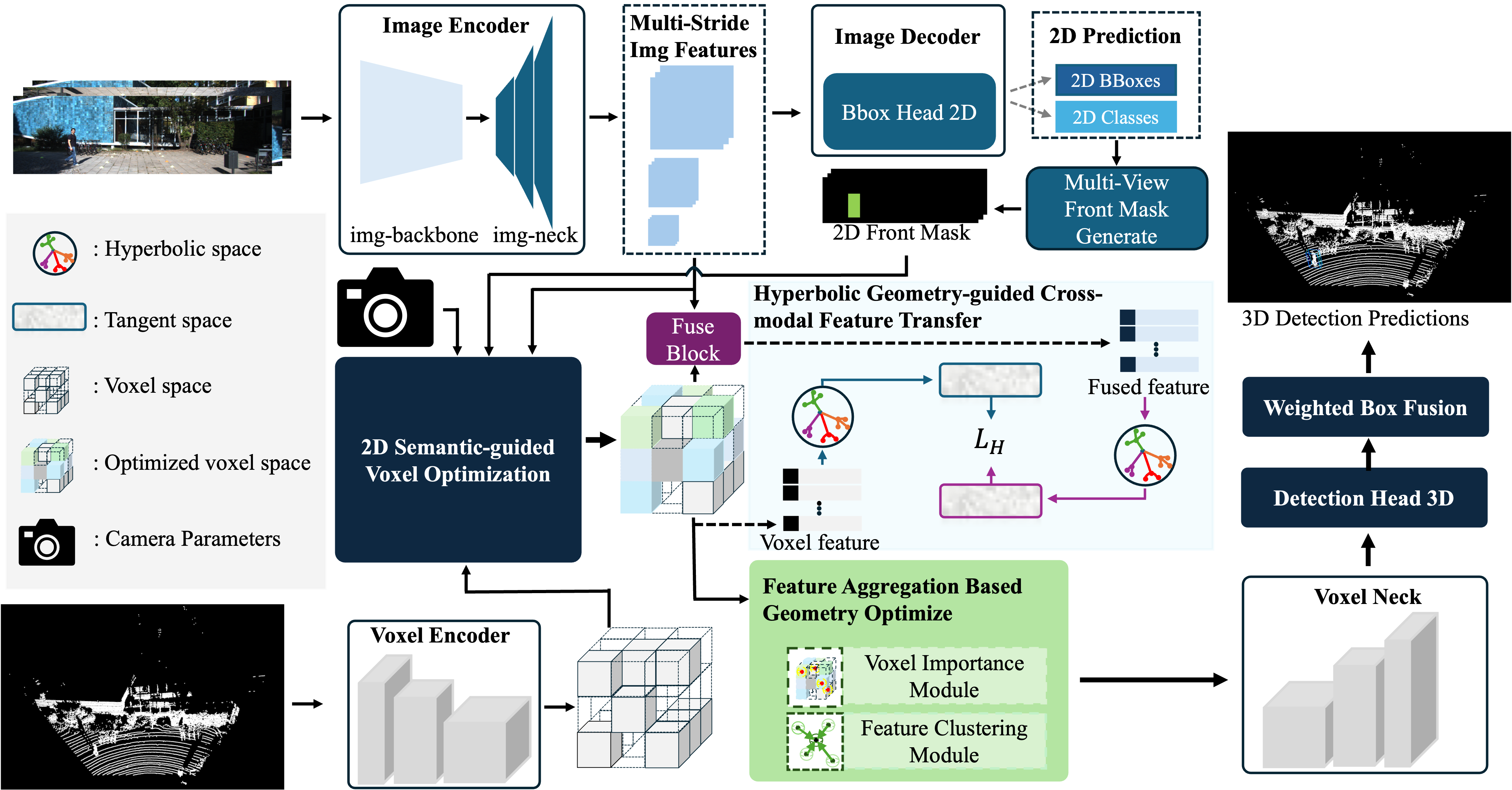}
\caption{The pipeline of our proposed HGC-Det is shown. Our detector uses images and point clouds as input, where the image input can be a surround view or a multi-view image. The point cloud branch uses voxelized point clouds as input. HGC-Det uses the semantic results predicted by the 2D branch to optimize and guide the feature extraction of the 3D branch.}
\end{figure*}

As shown in Fig. 3, the proposed HGC-Det framework can be broadly divided into image and point branches. The image branch extracts multi-scale features and 2D bounding boxes from the input image. The point cloud branch extracts features from the raw point cloud and, via cross-modal distillation, transfers the fused image-point cloud features into the point cloud feature stream. The point cloud branch can be further divided into a Voxel Encoder, the SGVO module, the HFT module, the FAGO module, Voxel Neck, and a 3D detection head. 

\textbf{Image Branch:} The image branch network needs to extract semantic features and 2D semantic information from a given input image. Here, we use YOLOv11 as our 2D branch. The proposed framework use these multi-scale image features extracted by image backbone, along with the 2D bounding boxes detected by 2D detection head and their semantic labels, as the output of the image branch. 

\textbf{Voxel Encoder:} In the point branch, we refer to the setup of existing 3D object detectors and use a network composed of voxel-based sparse 3D convolution layers as the voxel backbone. Here, we implement a sparse version of the 3D RseNet34\cite{rukhovich2023tr3d} network architecture based on the spconv\cite{yan2018second} library. 

\textbf{Voxel Neck:} Considering the scale difference of the target in the detector across the entire scene in outdoor and indoor scenes, our detector needs an FPN-like structure to integrate the multi-scale features extracted from the 3D backbone. Following the experience of \cite{chen2023voxelnext}, we use interpolation sampling instead of sparse transposed 3D convolution operations for upsampling.

\textbf{Detection Head:} In this part of the structure, we refer to existing popular works and use the CenterPoint\cite{yin2021center} head as the basic architecture. We then design an additional heat-map prediction operation on top of this, and use the prediction results of this additional heatmap to filter the input feature voxels. This improves the final detection performance, and the lightweight heat-map prediction combined with the feature filtering process allows our detector to maintain overall efficiency.

\subsection{2D Semantic Guided Voxel Optimize Module}
Considering the inherent discreteness of point clouds, the non-empty region they occupy when mapped to a 2D image plane are relatively sparse, directly fusing 3D voxels and their corresponding image feature pixels under this condition would result in a significant waste of image domain features\cite{li2022voxel, song2023simple, li2023logonet, 10510171}. Therefore, before fusing the data from the two modalities, we first divide the 3D region into foreground and background regions using the results of 2D branch prediction, and then perform voxel densification on the foreground space while sparsifying the background voxels.

Given 2D branch's output set $\mathbb{B}_{2D}$ and multi-stride feature maps set $\mathbb{F}_{2D}$. The proposed SGVO component partitions the multi-view input image into foreground and background regions based on geometric cues from the 2D branch, and generates a corresponding foreground mask geometry denoted as $\mathbf{M}_{2D}$. Given 3D voxel branch's output multi-stride voxel feature spaces $\mathbb{V}_{t} = (\mathbf{C}_t, \mathbf{F}^{3D}_t)$, where $t = 1, 2,4, 8, 16, 32$. For each stride's set of voxels, we map its coordinates in the 3D coordinate system to the coordinates $\mathbb{V}^{2D}_{t}$ in the camera plane. Then, based on the generated foreground mask $\mathbf{M}_{2D}$, we can divide the voxels $\mathbb{V}^{2D}_{t}$ into foreground region $\mathbb{V}^{f}_{t}$ and background region $\mathbb{V}^{b}_{t}$. This process can be described as:
\begin{equation}
\begin{aligned}
\mathbb{V}^{f}_{t}  & = \mathbf{M}_{2D} \odot \mathbb{V}^{2d}_{t} \\
\mathbb{V}^{b}_{t} &= (\mathbf{E} - \mathbf{M}_{2D}) \odot \mathbb{V}^{2d}_{t}
\end{aligned}
\end{equation}
where $\mathbb{E}$ denote the unit diagonal matrix and $\odot$ represents the Hadamard product.

Based on the defined foreground and background regions, we perform voxel densification on the foreground regions and voxel sparsification on the background regions. For the background voxel set $\mathbb{V}^{b}_{t}$, we apply a discretization operation using the 3D pooling layer provided by the spconv library. The extent of discretization is governed by the stride of the pooling layer, which we treat as an adjustable hyperparameter denoted by $\sigma$.

For the foreground region segmented by the 2D semantic branch, we perform a densification operation. Specifically, for a given set of foreground voxel points $\mathbb{V}^f_t=(\mathbb{C}^{t}_{f},\mathbf{F}^{t}_{f})$, we expand from the center of each foreground voxel point to a surrounding $3 \times 3 \times 3$ 3D space, removing all duplicate voxels. In our implementation, we adopt the strategy proposed in\cite{ma2024cotr} and employ sparse 3D convolution operations to realize learnable voxel densification. Specifically, we utilize a convolution kernel size of $3 \times 3 \times 3$ with a stride of $1 \times 1 \times 1$. After densifying and discretizing the foreground and background regions respectively, we superimpose the two processed voxel spaces to obtain the optimized multi-scale voxel space $\mathbb{V}^{t}_{opt}=(\mathbb{C}^{t}_{voxel},\mathbf{F}^{t}_{voxel})$.

\subsection{Hyperbolic Geometry-Guided Cross-modal Feature Transfer}
\textbf{1) Background:} Hyperbolic space is a curved space. To achieve hyperbolic alignment, we need to map feature points from Euclidean space to hyperbolic space. However, considering that hyperbolic space does not support Euclidean operations, we need to map the hyperbolic space back to its corresponding tangent space to calculate the alignment loss. This process involves two steps: hyperbolic mapping and tangent space mapping.

Hyperbolic Mapping. In this paper, we use the poincare ball model \cite{nickel2017poincare} to represent the hyperbolic space. We define an n-dimensional poincare ball space with $k$ constraints as follows:
\begin{equation}
\mathit{P}^{n}_{k} = \{ x \in \mathit{R}^{n} : ||x|| < \frac{1}{|k|} \}
\end{equation}
where $|| \cdot ||$ denote Frobenius norm.

To map an eigenvector point $x \in \mathbf{R}^n$﻿ from the Euclidean space $\mathit{R}^n$﻿ to an n-dimensional poincaré ball $\mathit{P}^n_k$﻿, we use the hyperbolic mapping function shown below:
\begin{equation}
x_p =  \Gamma_{\mathit{P}}(x)
\begin{cases}
x, & if ||x|| \quad \leq \frac{1}{k} \\
\frac{1 - \epsilon}{|k|} \frac{x}{||x||}, & else
\end{cases}
\end{equation}

Tangent Space Mapping. The hyperbolic tangent space is a Euclidean space that approximates a hyperbolic space. Given a hyperbolic space, we can generate its corresponding tangent plane $T_z \mathit{P}^n_k$.

\begin{equation}
\begin{aligned}
z_P \oplus_k x_P  = & \frac{(1 + 2|k|<z_P, x_P> + |k|||x_P||^2)z_P}{1 + 2|k|<z_P, x_P> + |k|^2 ||z_P||^2 ||x_P||^2}  \\
& + \frac{(1 - |k| ||z_P||^2)x_P}{1 + 2 |k|<z_P, x_P> + |k|^2 ||z_P||^2||x_P||^2} \\
x_{T_g} = & \frac{2}{\sqrt{|k|\lambda_{k}(z_{P})}}\tanh^{-1} (\sqrt{|k|} ||- z_P \oplus_k x_P||) \\
&  \cdot \frac{-z_P \oplus_k x_P}{||-z_P \oplus_k x_P||}
\end{aligned}
\end{equation}

Here, $<, >$defines the vector inner product, and $x_{Tg} \in T_{z} \mathit{P}^n_k$﻿. For ease of understanding and analysis, we define $z_{P}=0$﻿ here.

\textbf{2) Hyperbolic Constraint Cross-Modal Distillation: } As a core component of our detector, this module is designed to efficiently transfer information such as texture and color from the image branch into the point branch. In our scheme, we refer to \cite{yan20222dpass, yuan2022x} to design a hyperbolic constraint fusion-then-distillation framework.

In our framework, we first map the dense voxel centers to the corresponding 2D feature map, thereby obtaining the set of 2D feature vectors $\mathbf{F}^{t}_{2d}$ for each voxel. The set of voxel feature vectors extracted from point cloud can be represented as $\mathbf{F}^{t}_{voxel}$. Through the operations in \cite{yan20222dpass}, we first obtain the fused feature set of 2D feature vectors and voxel feature vectors as described follow:
\begin{equation}
\begin{aligned}
\mathbf{F}^{t}_{fuse} &= MLP(concat(\mathbf{F}^{t}_{2d}, \mathbf{F}^{t}_{voxel})) \\
\mathbf{F}^{t}_{2d3d} &= \mathbf{F}^{t}_{2d} + sigmoid(MLP(\mathbf{F}^{t}_{fuse})) \circ \mathbf{F}^{t}_{fuse}
\end{aligned}
\end{equation}

where $sigmoid$ denotes the corresponding activation function for regularizing the values of tensor to the range of $[0 ,1]$, $\circ$ represents the corresponding Hadamard product.

To calculate the corresponding hyperbolic constraint distillation loss, we map the fused feature vectors $\mathbf{F}^{t}_{2d3d}$﻿ and $\mathbf{F}^{t}_{voxel}$﻿ to their respective hyperbolic and tangent spaces. During this mapping process, we first map the feature vector set to the hyperbolic space using Eq.6. to obtain the corresponding $\mathbf{F}^{\mathit{P}, t}_{2d3d}$ and $\mathbf{F}^{\mathit{P}, t}_{voxel}$﻿; then, we map the feature vector set from the hyperbolic space to the tangent space using Eq.7. to obtain $\mathbf{F}^{T_g, t}_{2d3d}$ and $\mathbf{F}^{T_g, t}_{voxel}$﻿. Finally, we calculate the association loss between the two feature vector sets in the hyperbolic space using the following formula:
\begin{equation}
L_{H}(\mathbf{F}^{T_g, t}_{2d3d}, \mathbf{F}^{T_g, t}_{voxel}) = \sum_{t} ||\mathbf{F}^{T_g, t}_{2d3d} - \mathbf{F}^{T_g, t}_{voxel}||
\end{equation}

\subsection{Feature Aggregation Based Geometry Optimization Module}

The SGVO component described above partitions the 3D feature voxels into foreground and background regions and applies densification and discretization operations accordingly. While this adaptive redistribution of voxel space enhances feature utilization, it may inadvertently compromise the geometric fidelity of the original voxel representation. To address this, the proposed FAGO component is tasked with recovering the lost geometric information, which it accomplishes through voxel-level center voting. Specifically, FAGO comprises two submodules: a Voxel Importance Module and a Feature Clustering-based Center Voting Module.

\textbf{1) Voxel Importance Module: }This subtask module directly maps the input voxel feature vectors $\mathbf{F}^{t}_{voxel}$ to the importance score $\mathbb{S}^{t}_{voxel}$ using a set of MLP networks. For the annotation of this subtask, we simply use 3D bounding boxes from existing datasets, referencing the approach in\cite{qi2020imvotenet, wang2022cagroup3d}. By determining whether each point cloud in the entire scene is inside a 3D bounding box, we can obtain the foreground region annotation $\mathbb{S}^{t}_{gt}$ for the entire scene. For this subtask, we use Focal Loss to calculate its loss function, which can be specifically described as:

\begin{equation}
\begin{aligned}
L_{s} (s^t_{voxel}, s^t_{gt}) &= \sum_{t} -\frac{1}{B} \sum^{B}_{u=1} \frac{1}{M}\sum^{M}_{i=1} \alpha(1-s^t_{i})log(s^t_{i}) \\
s^t_{i} &= 
\begin{cases}
s^t_{voxel}, & if s^t_{gt} = 1 \\
1 - s^t_{voxel}, & else
\end{cases}
\end{aligned}
\end{equation}
Where $s^t_{voxel} \in \mathbb{S}^t_{voxel}$, $s^t_{gt} \in \mathbb{S}^t_{gt}$﻿, $B$ represents the batch size of the input data, and $M$ represents the number of non-empty voxels in the input voxel feature space. Based on the obtained importance prediction scores $\mathbb{S}_{voxel}$, we perform a second filtering operation on the entire input feature voxel space to retain the voxel points most likely to fall on the object surface. Here, we simply retain the 512 voxel set $\mathbb{V}^t_{filter} = (\mathbf{C}^t_{filter}, \mathbf{F}^t_{filter})$ with the highest scores in the entire space.

\textbf{2) Feature Clustering based Center Voting Module: } On the voxel set $\mathbb{V}^t_{filter}$ after filtering, our feature clustering subtask module performs center voting operations on these points from both the space and feature level.

In space level, we simply use a lightweight network composed of MLPs to predict the offset between each voxel and its corresponding geometric center. This lightweight network takes the feature vector of each voxel as input and predicts the offset of that voxel from the geometric center on the $x$, $y$, and $z$ axes, respectively. The loss function for this part is simply calculated using the $L2$ loss function, which can be expressed as:
\begin{equation}
\begin{aligned}
L_{ctr} &=\sum_{t} ||c^{t}_{p} - c^{t}_{gt}||^{2} \\
c^{t}_p &= c^{t}_i + idx \\
& c^{t}_i \in \mathbf{C}_{filter} 
\end{aligned}
\end{equation}
where $idx$ represents the prediction offset of the MLP network.

In the feature level, we calculate the feature clustering loss function for each voxel, using an unsupervised learning approach to bring similar parts of these voxels closer together in the feature dimension. We perform this clustering task using the Triplet loss function. For each given voxel $v^{i}_{filter} \in \mathbb{V}_{filter}$, we simply use the next voxel closest to it in Euclidean space, $v^{j}_{filter} \in \mathbb{V}_{filter}$, as its positive sample, and the voxel furthest from it in Euclidean space, $v^{t}_{filter}$, as its negative sample. The specific calculation formula can be expressed as:
\begin{equation}
L_{cluster} (\mathbf{F}_{filter}) = max(||f_i - f_j||^{2} - ||f_i - f_t||^{2} + margin, 0)
\end{equation}

After completing these two subtask,  we get a final voted voxel space $\mathbb{V}^{t}_{c}$. We then perform residual addition with the initially input feature voxel space $\mathbb{V}^t_{opt}$ to obtain the final output feature voxel space. This feature space is then fed into the corresponding voxel neck and detection head module.

\subsection{Objective Function}
In general, our detection head largely follows the design pattern of the CenterPoint. The classification loss and segmentation loss are represented by $L_{cls}$ and $L_{reg}$, respectively. From an implementation standpoint, the additional heatmap prediction can be regarded as a classification task; accordingly, its loss function follows the formulation of the center head classification loss. Beyond these standard loss terms, our overall objective incorporates a hyperbolic constraint for cross-modal distillation, denoted as $L_{H}$, along with three supplementary loss terms—$L_{s}$, $L_{\text{ctr}}$, and $L_{\text{cluster}}$—introduced by the two sub-task modules. Following the CenterPoint paradigm, the weights for $L_{\text{cls}}$ and $L_{\text{reg}}$ are set to $1.0$ and $2.0$, respectively, and the additional heatmap loss is equally weighted at $1.0$. The cross-modal distillation loss and the losses associated with the sub-tasks are further modulated by hyper-parameters $\eta_1$, $\eta_2$, $\eta_3$, and $\eta_4$, respectively. In summary, the overall objective function can be expressed as:

\begin{equation}
\begin{aligned}
L_{all} =& L_{cls} + L_{het} + 2L_{reg} + \eta_1 L_H  \\
 &+ \eta_2 L_s + \eta_3 L_{ctr} + \eta_4 L_{cluster}
\end{aligned}
\end{equation}

\section{Experiment}
This section presents a comprehensive performance evaluation of our HGC-Det. We first describe the datasets and training settings used in our experiments. Next, we report detailed performance metrics of our method on the selected datasets. Finally, we present the results of the ablation study. In addition, we analyze floating-point operations and provide visualizations of indoor and outdoor scenes; due to the submission length requirements of TMM, \textbf{these supplementary analyses are included in the supplementary materials.}

\subsection{Experiment Setting}
\subsubsection{\textbf{Dataset Setting}}
We selected representative datasets for both indoor and outdoor scenes, comprising KITTI and nuScenes for outdoor environments, and SUN RGB-D and ARKitScenes for indoor settings. For each dataset, we adopted the officially prescribed evaluation metrics and data split strategy. \textbf{Further details on metrics and dataset partitioning are provided in the supplementary materials.}

\subsubsection{\textbf{Training Setting}}

The proposed method uses AdamW as the optimizer. For the outdoor scene dataset, the initial learning rate is set to 0.0015, and the learning rate scheduling strategy is a one-cycle schedule. For indoor scenes, the proposed HGC-Det follow the practice of most methods\cite{zhu2024spgroup3d, kolodiazhnyi2025unidet3d, shen2023v, rukhovich2023tr3d} and use a simple multi-step LR schedule. It is worth noting that our method uses a pre-trained yolo11x.pt model for the 2D branch. Therefore, the 2D branch of our detector is in a freeze state regardless of the dataset. In the model architecture proposed in this paper, the hyper-parameter settings that remain unchanged are consistent with those in\cite{yin2021center}. Regarding the hyper-parameters introduced in the modules proposed in this paper, in the Image-Semantic-Guided Voxel Distribution Optimization Module, we set the max-pooling kernel size and stride during the background discretization process to $[2,2,2]$. In the subtask optimization module, the hyper-parameters $\eta_1$, $\eta_2$, $\eta_3$, and $\eta_4$ of the several additional loss functions introduced are set to $2.0$, $0.4$, $0.8$, and $0.8$, respectively. Further details on training setting are provided in the supplementary materials.

\subsection{Performance Analysis For Indoor Scenarios}

\begin{table}[!t]
\caption{Performance comparison analysis on the SUNRGBD dataset. Results marked with * indicate performance after preprocessing the point backbone using the pretrain method proposed in point gcc\cite{fan2024point}. The official paper does not publish the performance of TR3D+FF; here we use local experimental results.}
\scriptsize
\centering
\begin{tabular}{llllll}
\toprule[1.2pt]
Methods & map@25 & mar@25 & map@50 & mar@50  & Inference \\ \midrule[1.2pt]
TR3D\cite{rukhovich2023tr3d} & 65.56& 96.61 & 48.18 & 73.69 & 25.3 ms \\
RBGNet\cite{wang2022rbgnet} & 64.1 & - & 47.2 & -  & -\\
OneForAll\cite{wang2024one} & 65.0 & - & 51.3 & - & -\\
3Det-Mamba\cite{li20243det} & 61.3 & - & 42.2 & - & - \\
TR3D+FF\cite{rukhovich2023tr3d}  & 69.4 & 97.33 & 53.4  & 76.33 & 42.3 ms \\
TR3D+FF * \cite{rukhovich2023tr3d}& 69.7 & 98.22 & 54.0  & 78.32 & 42.3 ms \\
CAGroup \cite{wang2022cagroup3d}& 66.8  & 97.32 & 50.2  & 78.33 & 20.1 ms \\
CAGroup *\cite{wang2022cagroup3d}& 69.11 & 97.66 & 53.42 & 79.12 & 20.1 ms  \\
VDETR\cite{shen2023v} & 68.01 & 96.33 & 51.1  & 77.44 & -  \\
SPGroup3D\cite{zhu2024spgroup3d} & 65.40 & -      & 47.12 & -      & -\\
Ours & 69.13 & 97.35 & 53.64 & 79.35 & 35.2 ms \\
Ours *   & 70.12 & 97.92 & 54.52 & 79.35 & 35.2 ms \\ \bottomrule[1.2pt]
\end{tabular}
\end{table}

\begin{table}[!t]
\caption{The comparative experiments were conducted on the Arkitscenes dataset, using the same * annotation and performance pre-trained with the point gcc method\cite{fan2024point}. The performance of TR3D and TR3D+FF in this set of experiments is based on local test results from the official code. }
\scriptsize
\begin{tabular}{llllll}
\toprule[1.2pt]
Methods & map@25 & mar@25 & map@50 & mar@50  & Inference \\ \midrule[1.2pt]
UniDet3D\cite{kolodiazhnyi2025unidet3d} & 60.01 & 85.16 & 46.83 & 66.06 & 75.5 ms\\
TR3D\cite{rukhovich2023tr3d} & 56.32 & 82.14 & 42.13 & 61.13 & 28.3 ms\\
TR3D+FF\cite{rukhovich2023tr3d} & 58.92 & 83.95 & 44.12 & 63.64 & 47.4 ms \\
CubeRCNN\cite{li2024unimode} & 47.32 & -  & 19.26 & -  & 112.4 ms\\
3DiffTection\cite{xu20243difftection} & 47.32 & - & 20.30 & -  & 151.6 ms\\
Ours & 61.42 & 86.66 & 47.39 & 66.02  & 45.5 ms \\
Ours *   & 61.83 & 87.42 & 47.92 & 66.82 & 45.5 ms \\ \bottomrule[1.2pt]
\end{tabular}
\end{table}

To verify the performance of our method on indoor datasets, we conducted experiments on the indoor 3D object detection datasets SUNRGBD and ARKitScenes. The experimental results are shown in Tables  \uppercase\expandafter{\romannumeral1} and  \uppercase\expandafter{\romannumeral2}. Table \uppercase\expandafter{\romannumeral1} shows the comparative experiments on the SUNRGBD dataset, and Table \uppercase\expandafter{\romannumeral2} shows the experimental results on the ARKitScenes dataset. In the table, the results marked with * represent the performance after pre-training using the point-gcc method. The performance of TR3D+FF+Point GCC is based on experiments conducted locally using the official code. While the official paper on TR3D and TR3D+FF on the ArkitScenes dataset does not provide performance data, the results there are also from local experiments.

As shown in Table \uppercase\expandafter{\romannumeral1} , when our detector backbone is pre-trained without using the method proposed in\cite{fan2024point}, our method achieves a 0.24-point improvement in mAP 50 compared to the TR3D-FF version. When combined with the corresponding backbone pre-training method, our method achieves a 0.84-point performance improvement in mAP 50, and a 0.52-point improvement compared to the TR3D-FF+Point-GCC version. Regarding computational resource consumption, our method consumes 0.0352 seconds per frame on a single RTX 4090 machine, compared to 0.0253 seconds per frame for the original TR3D and 0.0423 seconds per frame for TR3D-FF.

As shown in the Table \uppercase\expandafter{\romannumeral2}, our method exhibits great performance advantages over the current work UniDet3D. Our method achieves performance improvements of 1.41 and 0.56 mAP@25 and mAP@50, respectively. This performance improvement becomes even more pronounced when our method is combined with the point-gcc backbone pre-training method. Furthermore, when performing inference on the ARKitSences dataset, the single scenes inference computation resource cost of the UniDet3D method is 0.0755 seconds, while our method's inference time is only 0.0455 seconds. This demonstrates that our method has a significant performance advantage over UniDet3D in both accuracy and time.

Experimental results on two indoor datasets demonstrate that our method exhibits significant performance advantages in both accuracy metrics and computational resource consumption for 3D object detection in indoor scenes. This demonstrates the effectiveness of our method.

\subsection{Performance Analysis For Outdoor Scenarios}

\begin{table*}[!t]
\vspace{-2.0em}
\caption{Performance Comparison on dataset KITTI. 'L' is the LiDAR and 'C' denotes the camera.}
\centering
\begin{tabular}{lllllllllll}
\toprule[1.2pt]
\multirow{2}{*}{Method} & \multirow{2}{*}{Modality} & \multicolumn{3}{c}{Car 3D AP}& \multicolumn{3}{c}{Pedestrian 3D AP}& \multicolumn{3}{c}{Cyclist 3D AP}\\
 & & \multicolumn{1}{l}{Mod} & \multicolumn{1}{l}{Easy} & Hard  & \multicolumn{1}{l}{Mod} & \multicolumn{1}{l}{Easy} & Hard  & \multicolumn{1}{l}{Mod} & \multicolumn{1}{l}{Easy} & \multicolumn{1}{l}{Hard} \\\midrule[1.2pt]
SECOND\cite{yan2018second} & L & 78.62 & 88.61 & 77.22 & 52.98 & 56.55 & 47.73 & 67.15 & 80.58 & 63.10 \\
PointPillars\cite{lang2019pointpillars} & L & 77.28 & 86.46 & 74.65 & 52.29 & 57.57 & 47.90 & 62.28 & 80.05 & 59.70 \\
PV-RCNN\cite{shi2020pv} & L & 84.36 & 92.10 & 82.48 & 56.67 & 64.26 & 51.91 & 71.95 & 88.88 & 66.78 \\
Voxel-RCNN\cite{wang2022voxel} & L & 85.29 & 92.38 & 82.86 & - & - & - & - & - & - \\
SE-SSD\cite{zheng2021se} & L & 86.25 & 90.21 & 79.22 & - & - & - & - & - & - \\
PDV\cite{hu2022point} & L & 92.56 & 85.29 & 83.05 & 60.80 & 66.90 & 55.85 & 74.23 & 92.72 & 69.60 \\
OneForAll\cite{wang2024one} & L & 84.2 & 92.8 & 82.3 & - & - & - & - & - & -  \\
EPNet\cite{huang2020epnet} & L+C & 78.65 & 88.76 & 78.32 & 59.29 & 66.74 & 54.82 & 65.60 & 83.88 & 62.70 \\
LoGoNet\cite{li2023logonet} & L+C & 85.04 & 92.04 & 84.31 & 63.72 & 70.20 & 59.46 & 75.35 & 91.74 & 72.42\\
VoxelNextFusion\cite{song2023simple} & L+C & 86.98 & 92.78 & 84.59 & - & - & - & - & - & - \\
FocalsConv\cite{chen2022focal} & L+C & 85.32 & 92.26 & 82.95 & - & - & - & - & - & - \\
CAT-Det\cite{zhang2022cat} & L+C & 81.46 & 90.12 & 79.15 & 66.35 & 74.08 & 58.92 & 72.82 & 87.64 & 68.20 \\
VFF\cite{li2022voxel} & L+C & 85.51 & 92.31 & 82.92 & 65.11 & 73.26 & 60.03 & 73.12 & 89.40 & 69.68 \\
MENet\cite{10510171} & L+C & 78.82 & 89.14 & 66.33 & 74.79 &  59.80 & 66.27 & 85.04 & 62.73 \\
Ours & L+C & 85.21 & 92.98 & 83.04 & 68.42 & 74.99 & 63.34 & 75.41 & 92.15 & 73.11 \\
\bottomrule[1.2pt]
\end{tabular}
\end{table*}

\begin{table*}[!t]
\centering
\caption{Performance Comparison on dataset Nuscenes. 'L' is the LiDAR and 'C' denotes the camera. 'C.V.', 'T.L.', 'B.R.', 'M.T.', 'Ped.' and 'T.C.' indicate the construction vehicle, trailer, barrier, motorcycle, pedestrian, and traffic cone, respectively. '( )' indicate that current results reproduced from our local environment with official released code and pre-trained weight.}
\begin{tabular}{llllllllllllll}
\toprule[1.2pt]
Method & Modality & mAP & NDS & Car & Truck & C.V. & Bus & T.L. & B.R. & M.T. & Bike & Ped. & T. C. \\ \midrule[1.2pt]
UniTR\cite{wang2023unitr} & L & 70.9 & 74.5 & 87.9 & 60.2 & 39.2 & 72.2 & 65.1 & 76.8 & 75.8 & 52.2 & 89.4 & 89.7 \\
(VoxelNext\cite{chen2023voxelnext}) & L & 60.48 & 67.66 & 84.04 & 54.59 & 22.58 & 66.46 & 38.53 & 65.37 & 65.26 & 50.78 & 86.10 & 71.01 \\
(HEDnet\cite{zhang2023hednet}) & L & 66.97 & 71.09 & 87.55 & 62.29 & 27.64 & 77.51 & 46.93 & 69.25 & 75.05 & 59.67 & 87.57 & 76.19 \\ 
(BEVFusion\cite{liang2022bevfusion}) & L+C & 67.75 & 69.43 & - & - & - & - & - & - & - & - & - & - \\
SparseFusion\cite{xie2023sparsefusion} & L+C & 71.0 & 73.1 & 87.6 & 59.8 & 38.3 & 71.6 & 64.5 & 78.8 & 78.1 & 59.4 &90.5 & 87.5 \\
CMT\cite{yan2023cross} &L+C & 70.3  & 87.4 & 62.7 & 36.7 & 74.8 & 64.8 & 77.6 & 78.5 & 60.0 87.3 & 84.1 \\ 
(EA-LSS\cite{hu2023ea}) & L+C & 71.2 & 73.1 &  85.2 & 62.1 & 71.7 & 64.1 & 38.9 & 86.3 & 80.9 & 61.6 & 86.2 & 79.5 \\
SparseLIFs\cite{zhang2024sparselif} & L+C & 73.4 & 76.5 & 90.7 & 72.3 & 82.0 & 46.3 & 33.1 & 90.8 & 83.9  & 76.9 & 83.0 & 75.0 \\
(IS-Fusion\cite{yin2024fusion}) & L+C & 72.8 & 74.0 & 88.3 & 62.7 & 38.4 & 74.9 & 67.3 & 78.1 & 82.4 & 59.5 & 89.3 & 89.2 \\ 
MV2DFusion\cite{wang2025mv2dfusion} & L+C & 73.9 & 75.4 & - & - & - & - & - & - & - & - & - & - \\
Ours & L+C & 73.4 & 76.7 & 88.2 & 63.1 & 38.6 & 75.1 & 67.4 & 78.2 & 83.2 & 60.2 & 90.3 & 90.2 \\
\bottomrule[1.2pt]
\end{tabular}
\end{table*}

To verify the performance of our method on outdoor datasets, we conducted experiments on the KITTI and Nuscenes datasets. The experimental results are shown in Tables  \uppercase\expandafter{\romannumeral3} and  \uppercase\expandafter{\romannumeral4}. Table  \uppercase\expandafter{\romannumeral3} shows the results of the experiments on the KITTI validation dataset, and Table  \uppercase\expandafter{\romannumeral4} shows the results of the experiments on the Nuscenes validation dataset. Considering that most papers do not release detailed performance data on the Nuscenes validation set, we reproduce the detailed performance data on the validation set using the official code and pre-trained weights for open-source code and weight methods. These results are marked in parentheses in the table.

Experimental results on KITTI show that our method has a significant advantage in accuracy and performance compared to current point-input-only methods. However, when compared to current state-of-the-art multimodal methods on the same dataset, our method still lags behind in accuracy for the Car category. On the Pedestrain and Cyclist categories, our method demonstrates a clear advantage in accuracy. We believe this advantage stems from our proposed voxel distribution optimization module. As mentioned in\cite{10510171, chen2022dsgn++}, small targets such as pedestrian and cyclist occupy more space in the front-view. Therefore, our image semantic guided voxel distribution module effectively increases the actual voxel count of these objects. This also explains why our method achieves less performance improvement on the Car category compared to other current state-of-the-art methods. The reason why our method performs worse than the current state-of-the-art (SOTA) in the car category may also stem from incomplete annotations in the KITTI dataset itself. These annotation-level incompletenesses in the KITTI dataset are described in more detail in the Section \uppercase\expandafter{\romannumeral4}.E.

Experimental results on the Nuscenes dataset demonstrate a significant performance advantage over pure point cloud methods. However, when compared to state-of-the-art multimodal methods, our approach still suffers from some performance degradation in NDS and mAP metrics. We attribute this to the fact that Nuscenes data is acquired using a single 32-line LiDAR, resulting in very low point cloud density, while each scene in this dataset typically contains a large number of targets. These two factors contribute to the performance difference between our method and current lift-splat-shot-based methods. However, when comparing performance with lift-splat-shoot based methods, we can observe that our method has relative advantages in several metrics, including learnable parameters, float-point operations, and frequency per second. For details, please refer to Section \uppercase\expandafter{\romannumeral4}.D of this chapter.

In summary, our method achieves very promising results on outdoor datasets, although the performance gains are not as impressive as on indoor datasets. Nevertheless, the results on both datasets fully demonstrate the performance advantages of our method.

\subsection{Visualize Analysis}
In this subsection, we demonstrate the visualization effectiveness of our method on the small outdoor dataset KITTI. We visualize the intermediate results of the detector's 2D branch, 3D branch, and subtask modules. In the following sections, we will introduce and analyze these visualized experimental results point by point.

\subsection{Ablation Study}
\subsubsection{Effect of Each Component}
\begin{table}
\centering
\caption{Effect of Each Component. 'ISVDO' denote the image semantic guided voxel distribution optimize module, 'HCFT' denote the hyperbolic constraint fusion to single feature transfer module, 'FCGO' represent the feature clustering based geometry optimization module.}
\begin{tabular}{ccccc}
\toprule[1.2pt]
ISVDO & HCFT & FCGO & map@25 & map@50 \\   \midrule[1.2pt]
 & & & 65.56 & 48.18 \\
\checkmark & & & 66.12 & 48.65 \\
\checkmark & \checkmark & & 67.82 & 51.13 \\
\checkmark & \checkmark & \checkmark & 69.13 & 53.64 \\
\bottomrule[1.2pt]
\end{tabular}
\end{table}
First, we validate the ablation experiments of the three modules proposed in this paper. In this set of experiments, we used the SUNRGBD dataset in an indoor setting. As can be observed from the experimental results in Table \uppercase\expandafter{\romannumeral5}, each component of our method can resulting in some performance improvemen, when all components are integrated together, the overall performance improvement of the method is particularly significant.

\subsubsection{Effect of Background Discretization Ratio}
\begin{figure}
\includegraphics[width=3.5in]{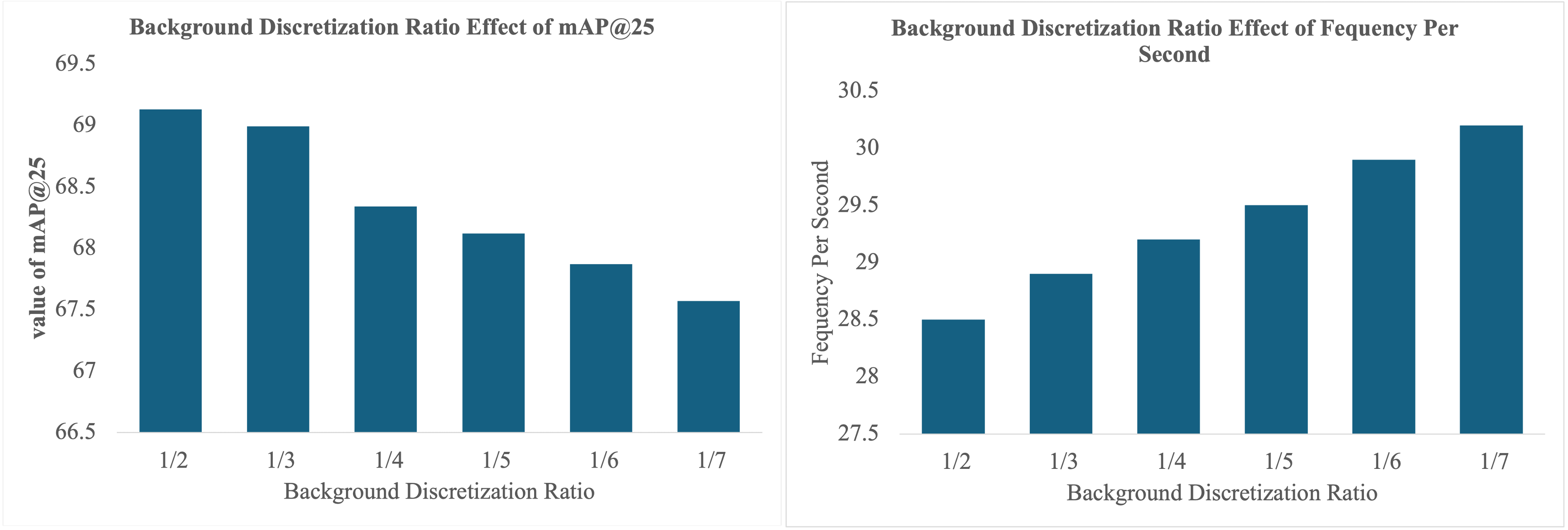}
\caption{Analysis of background discretization ratio's effect to detector's mean average precision with IoU threshold 0.25 and frequency per second. }
\end{figure}
In this part of the experiment, we verify the impact of the background region discretization degree on the overall performance of our method. Considering ease of engineering implementation, this ratio setting corresponds one-to-one with the kernel size and stride of the sparse max-pooling layer. Therefore, this ratio $\sigma$ can be set to $\frac{1}{2}, \frac{1}{3}, \frac{1}{4}, \frac{1}{5}, \frac{1}{6}, \frac{1}{7}$. From the experimental results in Fig. 4, it can be observed that as the ratio decreases during discretization, the FPS of the model increase, but the magnitude of this decrease is unacceptable compared to the resulting performance degradation. Our experimental results show that when the ratio is set to $\frac{1}{2}$, the balance between overall accuracy and performance of the model is optimal.

\subsubsection{Comparison of Direct Feature Distillation and Hyperbolic-Constraint Feature Distillation}
\begin{table}[!t]
\centering
\caption{Direct Feature Distillation Performance and Hyperbolic-Constraint Feature Distillation}
\begin{tabular}{ccc}
\toprule[1.2pt]
Method & map@25 & map@50  \\   \midrule[1.2pt]
Direct Feature Distillation & 68.13 & 52.86 \\
Hyperbolic-Constraint Distillation & 69.13 & 53.64 \\
\bottomrule[1.2pt]
\end{tabular}
\end{table}

Table \uppercase\expandafter{\romannumeral6} shows a comparison of the results using the direct feature distillation shown in\cite{yan20222dpass} and the experimental results using our proposed hyperbolic-constraint feature distillation. It can be seen that distilling the features of both modalities from the original space to the hyperbolic-constraint space before distillation yields better cross-modal knowledge transfer results.

\subsubsection{Effect of Hyper-parameter Setting}
\begin{figure}
\includegraphics[width=3.5in]{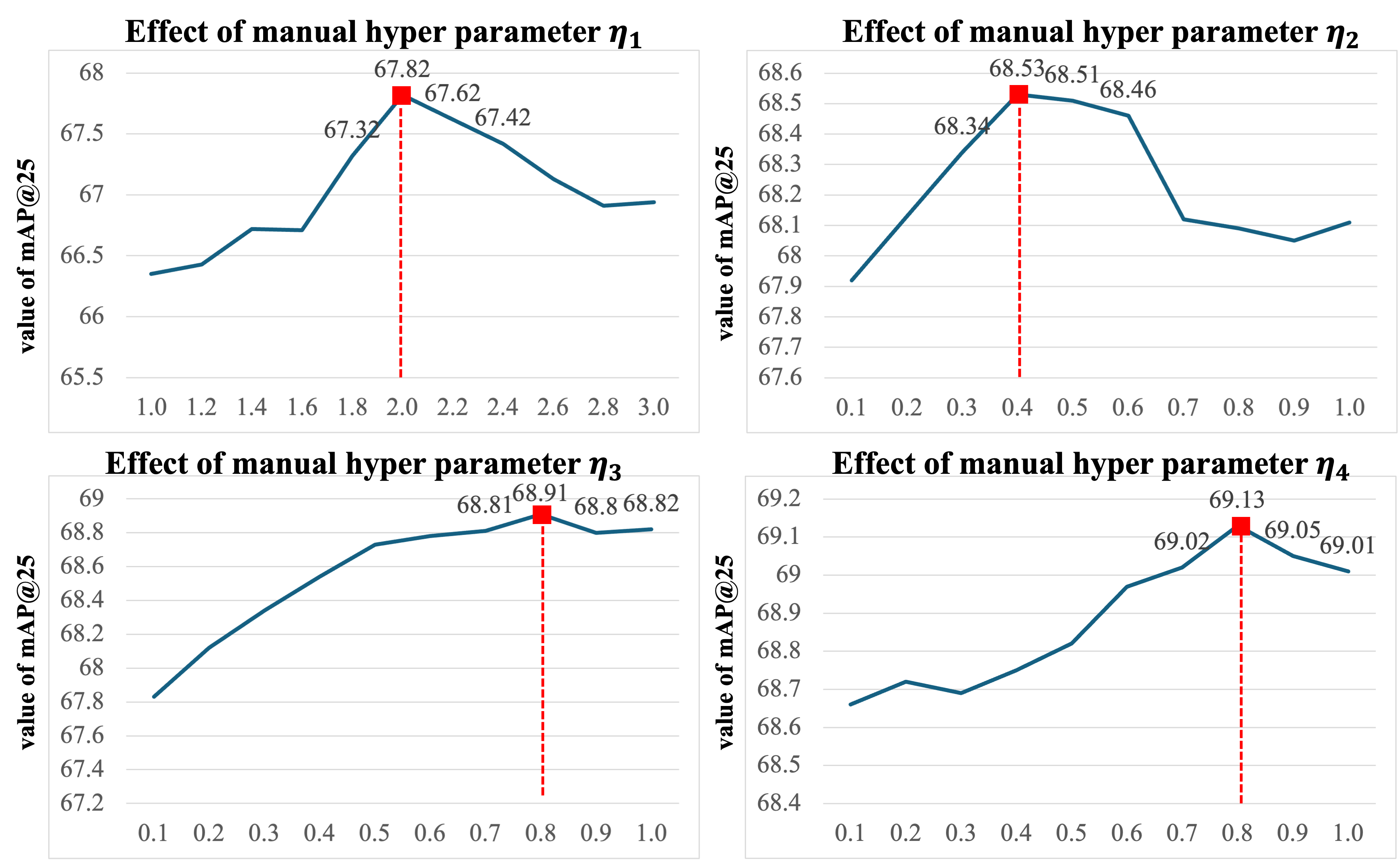}
\caption{Analysis of manual setting hyper-parameter's effect to detector's mean average precision with IoU threshold 0.25. We mark the optimal performance with red squares and use red dashed lines to indicate the corresponding x-axis values. For better visualization, the values of three points near the optimal on the figure also be displayed. }
\end{figure}
Fig.5. shows our experimental results for each parameter in the objective function. We continued to use the SUNRGBD dataset for this experiment. For the results of $\eta_1$, $\eta_2$, $\eta_3$, and $\eta_4$, we explored the optimal result by adjusting the parameters one by one. In this exploration mode, when exploring the current parameter, the parameters preceding it remained at their optimal values, while the parameters following it were all set to $0$. As shown in Fig.5. , when we set these parameters to $\eta_1=1.0$, $\eta_2=0.4$, $\eta_3=0.8$, and $\eta_4=0.8$ respectively, the overall performance of the detector achieved optimal values.

\section{Conclusions}

This proposed a hyperbolic guided cross-modal distillation method for multimodal 3D object detection. The proposed method comprised three core components: a 2D semantic-guided voxel optimization component, a hyperbolic geometry-guided cross-modal feature transfer component, and a feature aggregation based geometry optimization component. The SGVO component adaptively refined the spatial representation of the 3D branch by leveraging semantic cues from the image branch, thereby mitigating the issue of inadequate representation fusion. The HFT component exploited the intrinsic geometric properties of hyperbolic space to alleviate semantic loss during the fusion of high-dimensional image features and low-dimensional point cloud features. Finally, the FAGO compensated for potential spatial feature degradation introduced by the 2D semantic-guided voxel optimization component. Extensive experiments on indoor datasets (SUN RGB-D, ARKitScenes) and outdoor datasets (KITTI, nuScenes) demonstrated that our method achieved a better trade-off between detection accuracy and computational cost.

\bibliographystyle{IEEEtran}
\normalem
\bibliography{reference}

@article{yan2018second,
  title     = {Second: Sparsely embedded convolutional detection},
  author    = {Yan, Yan and Mao, Yuxing and Li, Bo},
  journal   = {Sensors},
  volume    = {18},
  number    = {10},
  pages     = {3337},
  year      = {2018},
  publisher = {Multidisciplinary Digital Publishing Institute}
}

@article{dong2025global,
  title={From global to hybrid: a review of supervised deep learning for 2-D image feature representation},
  author={Dong, Xinyu and Wang, Qi and Deng, Hongyu and Yang, Zhenguo and Ruan, Weijian and Liu, Wu and Lei, Liang and Wu, Xue and Tian, Youliang},
  journal={IEEE Transactions on Artificial Intelligence},
  volume={6},
  number={6},
  pages={1540--1560},
  year={2025},
  publisher={IEEE}
}

@inproceedings{yin2021center,
  title={Center-based 3d object detection and tracking},
  author={Yin, Tianwei and Zhou, Xingyi and Krahenbuhl, Philipp},
  booktitle={Proceedings of the IEEE/CVF conference on computer vision and pattern recognition},
  pages={11784--11793},
  year={2021}
}

@inproceedings{li2024unimode,
  title     = {Unimode: Unified monocular 3d object detection},
  author    = {Li, Zhuoling and Xu, Xiaogang and Lim, SerNam and Zhao, Hengshuang},
  booktitle = {Proceedings of the IEEE/CVF Conference on Computer Vision and Pattern Recognition},
  pages     = {16561--16570},
  year      = {2024}
}

@inproceedings{xu20243difftection,
  title     = {3difftection: 3d object detection with geometry-aware diffusion features},
  author    = {Xu, Chenfeng and Ling, Huan and Fidler, Sanja and Litany, Or},
  booktitle = {Proceedings of the IEEE/CVF Conference on Computer Vision and Pattern Recognition},
  pages     = {10617--10627},
  year      = {2024}
}

@inproceedings{zhu2024spgroup3d,
  title     = {Spgroup3d: Superpoint grouping network for indoor 3d object detection},
  author    = {Zhu, Yun and Hui, Le and Shen, Yaqi and Xie, Jin},
  booktitle = {Proceedings of the AAAI Conference on Artificial Intelligence},
  volume    = {38},
  number    = {7},
  pages     = {7811--7819},
  year      = {2024}
}

@inproceedings{kolodiazhnyi2025unidet3d,
  title={Unidet3d: Multi-dataset indoor 3d object detection},
  author={Kolodiazhnyi, Maksim and Vorontsova, Anna and Skripkin, Matvey and Rukhovich, Danila and Konushin, Anton},
  booktitle={Proceedings of the AAAI Conference on Artificial Intelligence},
  volume={39},
  number={4},
  pages={4365--4373},
  year={2025}
}

@inproceedings{qi2020imvotenet,
  title     = {Imvotenet: Boosting 3d object detection in point clouds with image votes},
  author    = {Qi, Charles R and Chen, Xinlei and Litany, Or and Guibas, Leonidas J},
  booktitle = {Proceedings of the IEEE/CVF conference on computer vision and pattern recognition},
  pages     = {4404--4413},
  year      = {2020}
}

@article{shen2023v,
  title   = {V-detr: Detr with vertex relative position encoding for 3d object detection},
  author  = {Shen, Yichao and Geng, Zigang and Yuan, Yuhui and Lin, Yutong and Liu, Ze and Wang, Chunyu and Hu, Han and Zheng, Nanning and Guo, Baining},
  journal = {arXiv preprint arXiv:2308.04409},
  year    = {2023}
}

@inproceedings{li20243det,
  title     = {3DET-Mamba: State Space Model for End-to-End 3D Object Detection},
  author    = {Li, Mingsheng and Yuan, Jiakang and Chen, Sijin and Zhang, Lin and Zhu, Anyu and Chen, Xin and Chen, Tao},
  booktitle = {Proceedings of the 38th International Conference on Neural Information Processing Systems},
  pages     = {47242--47260},
  year      = {2024}
}

@article{wang2022cagroup3d,
  title   = {Cagroup3d: Class-aware grouping for 3d object detection on point clouds},
  author  = {Wang, Haiyang and Ding, Lihe and Dong, Shaocong and Shi, Shaoshuai and Li, Aoxue and Li, Jianan and Li, Zhenguo and Wang, Liwei},
  journal = {Advances in neural information processing systems},
  volume  = {35},
  pages   = {29975--29988},
  year    = {2022}
}

@inproceedings{wang2022rbgnet,
  title     = {Rbgnet: Ray-based grouping for 3d object detection},
  author    = {Wang, Haiyang and Shi, Shaoshuai and Yang, Ze and Fang, Rongyao and Qian, Qi and Li, Hongsheng and Schiele, Bernt and Wang, Liwei},
  booktitle = {Proceedings of the IEEE/CVF Conference on Computer Vision and Pattern Recognition},
  pages     = {1110--1119},
  year      = {2022}
}

@inproceedings{rukhovich2023tr3d,
  title        = {Tr3d: Towards real-time indoor 3d object detection},
  author       = {Rukhovich, Danila and Vorontsova, Anna and Konushin, Anton},
  booktitle    = {2023 IEEE International Conference on Image Processing (ICIP)},
  pages        = {281--285},
  year         = {2023},
  organization = {IEEE}
}

@article{wang2025mv2dfusion,
  title     = {Mv2dfusion: Leveraging modality-specific object semantics for multi-modal 3d detection},
  author    = {Wang, Zitian and Huang, Zehao and Gao, Yulu and Wang, Naiyan and Liu, Si},
  journal   = {IEEE Transactions on Pattern Analysis and Machine Intelligence},
  year      = {2025},
  publisher = {IEEE}
}

@inproceedings{yin2024fusion,
  title     = {Is-fusion: Instance-scene collaborative fusion for multimodal 3d object detection},
  author    = {Yin, Junbo and Shen, Jianbing and Chen, Runnan and Li, Wei and Yang, Ruigang and Frossard, Pascal and Wang, Wenguan},
  booktitle = {Proceedings of the IEEE/CVF conference on computer vision and pattern recognition},
  pages     = {14905--14915},
  year      = {2024}
}

@article{hu2023ea,
  title   = {Ea-lss: Edge-aware lift-splat-shot framework for 3d bev object detection},
  author  = {Hu, Haotian and Wang, Fanyi and Su, Jingwen and Wang, Yaonong and Hu, Laifeng and Fang, Weiye and Xu, Jingwei and Zhang, Zhiwang},
  journal = {arXiv preprint arXiv:2303.17895},
  year    = {2023}
}

@article{yan2023cross,
  title     = {Cross modal transformer via coordinates encoding for 3d object dectection},
  author    = {Yan, Junjie and Liu, Yingfei and Sun, Jianjian and Jia, Fan and Li, Shuailin and Wang, Tiancai and Zhang, Xiangyu},
  journal   = {arXiv preprint arXiv:2301.01283},
  volume    = {2},
  number    = {3},
  pages     = {4},
  year      = {2023},
  publisher = {ArXiv}
}

@inproceedings{xie2023sparsefusion,
  title     = {Sparsefusion: Fusing multi-modal sparse representations for multi-sensor 3d object detection},
  author    = {Xie, Yichen and Xu, Chenfeng and Rakotosaona, Marie-Julie and Rim, Patrick and Tombari, Federico and Keutzer, Kurt and Tomizuka, Masayoshi and Zhan, Wei},
  booktitle = {Proceedings of the IEEE/CVF International Conference on Computer Vision},
  pages     = {17591--17602},
  year      = {2023}
}

@inproceedings{zhang2024sparselif,
  title={SparseLIF: High-performance sparse LiDAR-camera fusion for 3D object detection},
  author={Zhang, Hongcheng and Liang, Liu and Zeng, Pengxin and Song, Xiao and Wang, Zhe},
  booktitle={European Conference on Computer Vision},
  pages={109--128},
  year={2024},
  organization={Springer}
}

@article{liang2022bevfusion,
  title   = {Bevfusion: A simple and robust lidar-camera fusion framework},
  author  = {Liang, Tingting and Xie, Hongwei and Yu, Kaicheng and Xia, Zhongyu and Lin, Zhiwei and Wang, Yongtao and Tang, Tao and Wang, Bing and Tang, Zhi},
  journal = {Advances in Neural Information Processing Systems},
  volume  = {35},
  pages   = {10421--10434},
  year    = {2022}
}

@article{zhang2023hednet,
  title   = {Hednet: A hierarchical encoder-decoder network for 3d object detection in point clouds},
  author  = {Zhang, Gang and Junnan, Chen and Gao, Guohuan and Li, Jianmin and Hu, Xiaolin},
  journal = {Advances in Neural Information Processing Systems},
  volume  = {36},
  pages   = {53076--53089},
  year    = {2023}
}

@inproceedings{chen2023voxelnext,
  title     = {Voxelnext: Fully sparse voxelnet for 3d object detection and tracking},
  author    = {Chen, Yukang and Liu, Jianhui and Zhang, Xiangyu and Qi, Xiaojuan and Jia, Jiaya},
  booktitle = {Proceedings of the IEEE/CVF conference on computer vision and pattern recognition},
  pages     = {21674--21683},
  year      = {2023}
}

@inproceedings{wang2023unitr,
  title     = {Unitr: A unified and efficient multi-modal transformer for bird's-eye-view representation},
  author    = {Wang, Haiyang and Tang, Hao and Shi, Shaoshuai and Li, Aoxue and Li, Zhenguo and Schiele, Bernt and Wang, Liwei},
  booktitle = {Proceedings of the IEEE/CVF international conference on computer vision},
  pages     = {6792--6802},
  year      = {2023}
}

@inproceedings{li2022voxel,
  title     = {Voxel field fusion for 3d object detection},
  author    = {Li, Yanwei and Qi, Xiaojuan and Chen, Yukang and Wang, Liwei and Li, Zeming and Sun, Jian and Jia, Jiaya},
  booktitle = {Proceedings of the IEEE/CVF conference on computer vision and pattern recognition},
  pages     = {1120--1129},
  year      = {2022}
}

@inproceedings{zhang2022cat,
  title     = {Cat-det: Contrastively augmented transformer for multi-modal 3d object detection},
  author    = {Zhang, Yanan and Chen, Jiaxin and Huang, Di},
  booktitle = {Proceedings of the IEEE/CVF Conference on Computer Vision and Pattern Recognition},
  pages     = {908--917},
  year      = {2022}
}

@inproceedings{chen2022focal,
  title     = {Focal sparse convolutional networks for 3d object detection},
  author    = {Chen, Yukang and Li, Yanwei and Zhang, Xiangyu and Sun, Jian and Jia, Jiaya},
  booktitle = {Proceedings of the IEEE/CVF conference on computer vision and pattern recognition},
  pages     = {5428--5437},
  year      = {2022}
}

@article{song2023simple,
  title   = {A Simple, Unified and Effective Voxel Fusion Framework for Multi-Modal 3D Object Detection},
  author  = {Song, Z and Zhang, G and Xie, J and Liu, L and Jia, C and Xu, S VoxelNextFusion},
  journal = {IEEE Trans. Geosci. Remote Sens},
  volume  = {61},
  pages   = {5705412},
  year    = {2023}
}

@inproceedings{li2023logonet,
  title     = {Logonet: Towards accurate 3d object detection with local-to-global cross-modal fusion},
  author    = {Li, Xin and Ma, Tao and Hou, Yuenan and Shi, Botian and Yang, Yuchen and Liu, Youquan and Wu, Xingjiao and Chen, Qin and Li, Yikang and Qiao, Yu and others},
  booktitle = {Proceedings of the IEEE/CVF conference on computer vision and pattern recognition},
  pages     = {17524--17534},
  year      = {2023}
}

@inproceedings{huang2020epnet,
  title        = {Epnet: Enhancing point features with image semantics for 3d object detection},
  author       = {Huang, Tengteng and Liu, Zhe and Chen, Xiwu and Bai, Xiang},
  booktitle    = {European conference on computer vision},
  pages        = {35--52},
  year         = {2020},
  organization = {Springer}
}

@article{wang2024one,
  title   = {One for all: Multi-domain joint training for point cloud based 3d object detection},
  author  = {Wang, Zhenyu and Li, Ya-Li and Zhao, Hengshuang and Wang, Shengjin},
  journal = {Advances in Neural Information Processing Systems},
  volume  = {37},
  pages   = {56859--56877},
  year    = {2024}
}

@inproceedings{hu2022point,
  title     = {Point density-aware voxels for lidar 3d object detection},
  author    = {Hu, Jordan SK and Kuai, Tianshu and Waslander, Steven L},
  booktitle = {Proceedings of the IEEE/CVF conference on computer vision and pattern recognition},
  pages     = {8469--8478},
  year      = {2022}
}

@inproceedings{zheng2021se,
  title     = {SE-SSD: Self-ensembling single-stage object detector from point cloud},
  author    = {Zheng, Wu and Tang, Weiliang and Jiang, Li and Fu, Chi-Wing},
  booktitle = {Proceedings of the IEEE/CVF conference on computer vision and pattern recognition},
  pages     = {14494--14503},
  year      = {2021}
}

@article{wang2022voxel,
  title     = {Voxel-RCNN-complex: An effective 3-D point cloud object detector for complex traffic conditions},
  author    = {Wang, Hai and Chen, Zhiyu and Cai, Yingfeng and Chen, Long and Li, Yicheng and Sotelo, Miguel Angel and Li, Zhixiong},
  journal   = {IEEE Transactions on Instrumentation and Measurement},
  volume    = {71},
  pages     = {1--12},
  year      = {2022},
  publisher = {IEEE}
}

@inproceedings{shi2020pv,
  title     = {Pv-rcnn: Point-voxel feature set abstraction for 3d object detection},
  author    = {Shi, Shaoshuai and Guo, Chaoxu and Jiang, Li and Wang, Zhe and Shi, Jianping and Wang, Xiaogang and Li, Hongsheng},
  booktitle = {Proceedings of the IEEE/CVF conference on computer vision and pattern recognition},
  pages     = {10529--10538},
  year      = {2020}
}

@inproceedings{lang2019pointpillars,
  title     = {Pointpillars: Fast encoders for object detection from point clouds},
  author    = {Lang, Alex H and Vora, Sourabh and Caesar, Holger and Zhou, Lubing and Yang, Jiong and Beijbom, Oscar},
  booktitle = {Proceedings of the IEEE/CVF conference on computer vision and pattern recognition},
  pages     = {12697--12705},
  year      = {2019}
}

@inproceedings{fan2024point,
  title     = {Point-gcc: Universal self-supervised 3d scene pre-training via geometry-color contrast},
  author    = {Fan, Guofan and Qi, Zekun and Shi, Wenkai and Ma, Kaisheng},
  booktitle = {Proceedings of the 32nd ACM International Conference on Multimedia},
  pages     = {4709--4718},
  year      = {2024}
}

@ARTICLE{10659158,
  author={Sun, Tianfang and Zhang, Zhizhong and Tan, Xin and Peng, Yong and Qu, Yanyun and Xie, Yuan},
  journal={IEEE Transactions on Pattern Analysis and Machine Intelligence}, 
  title={Uni-to-Multi Modal Knowledge Distillation for Bidirectional LiDAR-Camera Semantic Segmentation}, 
  year={2024},
  volume={46},
  number={12},
  pages={11059-11072},
  doi={10.1109/TPAMI.2024.3451658}}

@inproceedings{yan20222dpass,
  title        = {2dpass: 2d priors assisted semantic segmentation on lidar point clouds},
  author       = {Yan, Xu and Gao, Jiantao and Zheng, Chaoda and Zheng, Chao and Zhang, Ruimao and Cui, Shuguang and Li, Zhen},
  booktitle    = {European conference on computer vision},
  pages        = {677--695},
  year         = {2022},
  organization = {Springer}
}

@inproceedings{yuan2022x,
  title     = {X-trans2cap: Cross-modal knowledge transfer using transformer for 3d dense captioning},
  author    = {Yuan, Zhihao and Yan, Xu and Liao, Yinghong and Guo, Yao and Li, Guanbin and Cui, Shuguang and Li, Zhen},
  booktitle = {Proceedings of the IEEE/CVF Conference on Computer Vision and Pattern Recognition},
  pages     = {8563--8573},
  year      = {2022}
}

@article{liu2021learning,
  title   = {Learning from 2d: Contrastive pixel-to-point knowledge transfer for 3d pretraining},
  author  = {Liu, Yueh-Cheng and Huang, Yu-Kai and Chiang, Hung-Yueh and Su, Hung-Ting and Liu, Zhe-Yu and Chen, Chin-Tang and Tseng, Ching-Yu and Hsu, Winston H},
  journal = {arXiv preprint arXiv:2104.04687},
  year    = {2021}
}

@article{xu2021image2point,
  title   = {Image2point: 3d point-cloud understanding with pretrained 2d convnets},
  author  = {Xu, Chenfeng and Yang, Shijia and Zhai, Bohan and Wu, Bichen and Yue, Xiangyu and Zhan, Wei and Vajda, Peter and Keutzer, Kurt and Tomizuka, Masayoshi},
  journal = {arXiv preprint arXiv:2106.04180},
  volume  = {3},
  year    = {2021}
}

@article{chen2022dsgn++,
  title={Dsgn++: Exploiting visual-spatial relation for stereo-based 3d detectors},
  author={Chen, Yilun and Huang, Shijia and Liu, Shu and Yu, Bei and Jia, Jiaya},
  journal={IEEE Transactions on Pattern Analysis and Machine Intelligence},
  volume={45},
  number={4},
  pages={4416--4429},
  year={2022},
  publisher={IEEE}
}

@article{10510171,
  author   = {Liu, Moyun and Chen, Youping and Xie, Jingming and Zhu, Yijie and Zhang, Yang and Yao, Lei and Bing, Zhenshan and Zhuang, Genghang and Huang, Kai and Zhou, Joey Tianyi},
  journal  = {IEEE Transactions on Intelligent Transportation Systems},
  title    = {MENet: Multi-Modal Mapping Enhancement Network for 3D Object Detection in Autonomous Driving},
  year     = {2024},
  volume   = {25},
  number   = {8},
  pages    = {9397-9410},
  doi      = {10.1109/TITS.2024.3387398}
}

@article{nickel2017poincare,
  title={Poincar{\'e} embeddings for learning hierarchical representations},
  author={Nickel, Maximillian and Kiela, Douwe},
  journal={Advances in neural information processing systems},
  volume={30},
  year={2017}
}

@inproceedings{sur2025hyperbolic,
  title     = {Hyperbolic Uncertainty-Aware Few-Shot Incremental Point Cloud Segmentation},
  author    = {Sur, Tanuj and Mukherjee, Samrat and Rahaman, Kaizer and Chaudhuri, Subhasis and Khan, Muhammad Haris and Banerjee, Biplab},
  booktitle = {Proceedings of the Computer Vision and Pattern Recognition Conference},
  pages     = {11810--11821},
  year      = {2025}
}

@article{liu2025hyperbolic,
  title   = {Hyperbolic Contrastive Learning for Hierarchical 3D Point Cloud Embedding},
  author  = {Liu, Yingjie and Zhang, Pengyu and He, Ziyao and Chen, Mingsong and Tang, Xuan and Wei, Xian},
  journal = {arXiv preprint arXiv:2501.02285},
  year    = {2025}
}

@inproceedings{li2025hyperbolic,
  title     = {Hyperbolic-constraint Point Cloud Reconstruction from Single RGB-D Images},
  author    = {Li, Wenrui and Yang, Zhe and Han, Wei and Man, Hengyu and Wang, Xingtao and Fan, Xiaopeng},
  booktitle = {Proceedings of the AAAI Conference on Artificial Intelligence},
  volume    = {39},
  number    = {5},
  pages     = {4959--4967},
  year      = {2025}
}

@article{hong2023curved,
  title     = {Curved geometric networks for visual anomaly recognition},
  author    = {Hong, Jie and Fang, Pengfei and Li, Weihao and Han, Junlin and Petersson, Lars and Harandi, Mehrtash},
  journal   = {IEEE transactions on neural networks and learning systems},
  year      = {2023},
  publisher = {IEEE}
}

@inproceedings{atigh2022hyperbolic,
  title     = {Hyperbolic image segmentation},
  author    = {Atigh, Mina Ghadimi and Schoep, Julian and Acar, Erman and Van Noord, Nanne and Mettes, Pascal},
  booktitle = {Proceedings of the IEEE/CVF conference on computer vision and pattern recognition},
  pages     = {4453--4462},
  year      = {2022}
}

@article{yang2025shmamba,
  title     = {Shmamba: Structured hyperbolic state space model for audio-visual question answering},
  author    = {Yang, Zhe and Li, Wenrui and Cheng, Guanghui},
  journal   = {IEEE Transactions on Audio, Speech and Language Processing},
  year      = {2025},
  publisher = {IEEE}
}

@article{fang2023hyperbolic,
  title   = {Hyperbolic geometry in computer vision: A survey},
  author  = {Fang, Pengfei and Harandi, Mehrtash and Le, Trung and Phung, Dinh},
  journal = {arXiv preprint arXiv:2304.10764},
  year    = {2023}
}

@article{pravallika2024deep,
  title     = {Deep learning frontiers in 3D object detection: a comprehensive review for autonomous driving},
  author    = {Pravallika, Ambati and Hashmi, Mohammad Farukh and Gupta, Aditya},
  journal   = {IEEE Access},
  year      = {2024},
  publisher = {IEEE}
}

@article{xu2025survey,
  title     = {A survey on occupancy perception for autonomous driving: The information fusion perspective},
  author    = {Xu, Huaiyuan and Chen, Junliang and Meng, Shiyu and Wang, Yi and Chau, Lap-Pui},
  journal   = {Information Fusion},
  volume    = {114},
  pages     = {102671},
  year      = {2025},
  publisher = {Elsevier}
}

@inproceedings{song2015sun,
  title     = {Sun rgb-d: A rgb-d scene understanding benchmark suite},
  author    = {Song, Shuran and Lichtenberg, Samuel P and Xiao, Jianxiong},
  booktitle = {Proceedings of the IEEE conference on computer vision and pattern recognition},
  pages     = {567--576},
  year      = {2015}
}

@article{baruch2021arkitscenes,
  title   = {Arkitscenes: A diverse real-world dataset for 3d indoor scene understanding using mobile rgb-d data},
  author  = {Baruch, Gilad and Chen, Zhuoyuan and Dehghan, Afshin and Dimry, Tal and Feigin, Yuri and Fu, Peter and Gebauer, Thomas and Joffe, Brandon and Kurz, Daniel and Schwartz, Arik and others},
  journal = {arXiv preprint arXiv:2111.08897},
  year    = {2021}
}

@article{hinton2015distilling,
  title   = {Distilling the knowledge in a neural network},
  author  = {Hinton, Geoffrey and Vinyals, Oriol and Dean, Jeff},
  journal = {arXiv preprint arXiv:1503.02531},
  year    = {2015}
}

@inproceedings{ma2024cotr,
  title={Cotr: Compact occupancy transformer for vision-based 3d occupancy prediction},
  author={Ma, Qihang and Tan, Xin and Qu, Yanyun and Ma, Lizhuang and Zhang, Zhizhong and Xie, Yuan},
  booktitle={Proceedings of the IEEE/CVF Conference on Computer Vision and Pattern Recognition},
  pages={19936--19945},
  year={2024}
}

@article{wu2025fully,
  title={Fully-connected transformer for multi-source image fusion},
  author={Wu, Xiao and Cao, Zi-Han and Huang, Ting-Zhu and Deng, Liang-Jian and Chanussot, Jocelyn and Vivone, Gemine},
  journal={IEEE Transactions on Pattern Analysis and Machine Intelligence},
  volume={47},
  number={3},
  pages={2071--2088},
  year={2025},
  publisher={IEEE}
}

@inproceedings{hou2025binarized,
  title={Binarized neural network for multi-spectral image fusion},
  author={Hou, Junming and Chen, Xiaoyu and Ran, Ran and Cong, Xiaofeng and Liu, Xinyang and You, Jian Wei and Deng, Liang-Jian},
  booktitle={Proceedings of the Computer Vision and Pattern Recognition Conference},
  pages={2236--2245},
  year={2025}
}

@article{wu2024inf,
  title={INF 3: Implicit neural feature fusion function for multispectral and hyperspectral image fusion},
  author={Wu, Ruo-Cheng and Deng, Shangqi and Ran, Ran and Dou, Hong-Xia and Deng, Liang-Jian},
  journal={IEEE Transactions on Computational Imaging},
  volume={10},
  pages={1547--1558},
  year={2024},
  publisher={IEEE}
}

@article{luo2024revisiting,
  title={Revisiting nonlocal self-similarity from continuous representation},
  author={Luo, Yisi and Zhao, Xile and Meng, Deyu},
  journal={IEEE Transactions on Pattern Analysis and Machine Intelligence},
  volume={47},
  number={1},
  pages={450--468},
  year={2024},
  publisher={IEEE}
}

@article{xu2025parameterized,
  title={Parameterized low-rank regularizer for high-dimensional visual data},
  author={Xu, Shuang and Zhao, Zixiang and Cao, Xiangyong and Peng, Jiangjun and Zhao, Xi-Le and Meng, Deyu and Zhang, Yulun and Timofte, Radu and Van Gool, Luc},
  journal={International Journal of Computer Vision},
  volume={133},
  number={12},
  pages={8546--8569},
  year={2025},
  publisher={Springer}
}

\vspace{-1.5cm}
\begin{IEEEbiography}[{\includegraphics[width=1in,height=1.25in,clip,keepaspectratio]{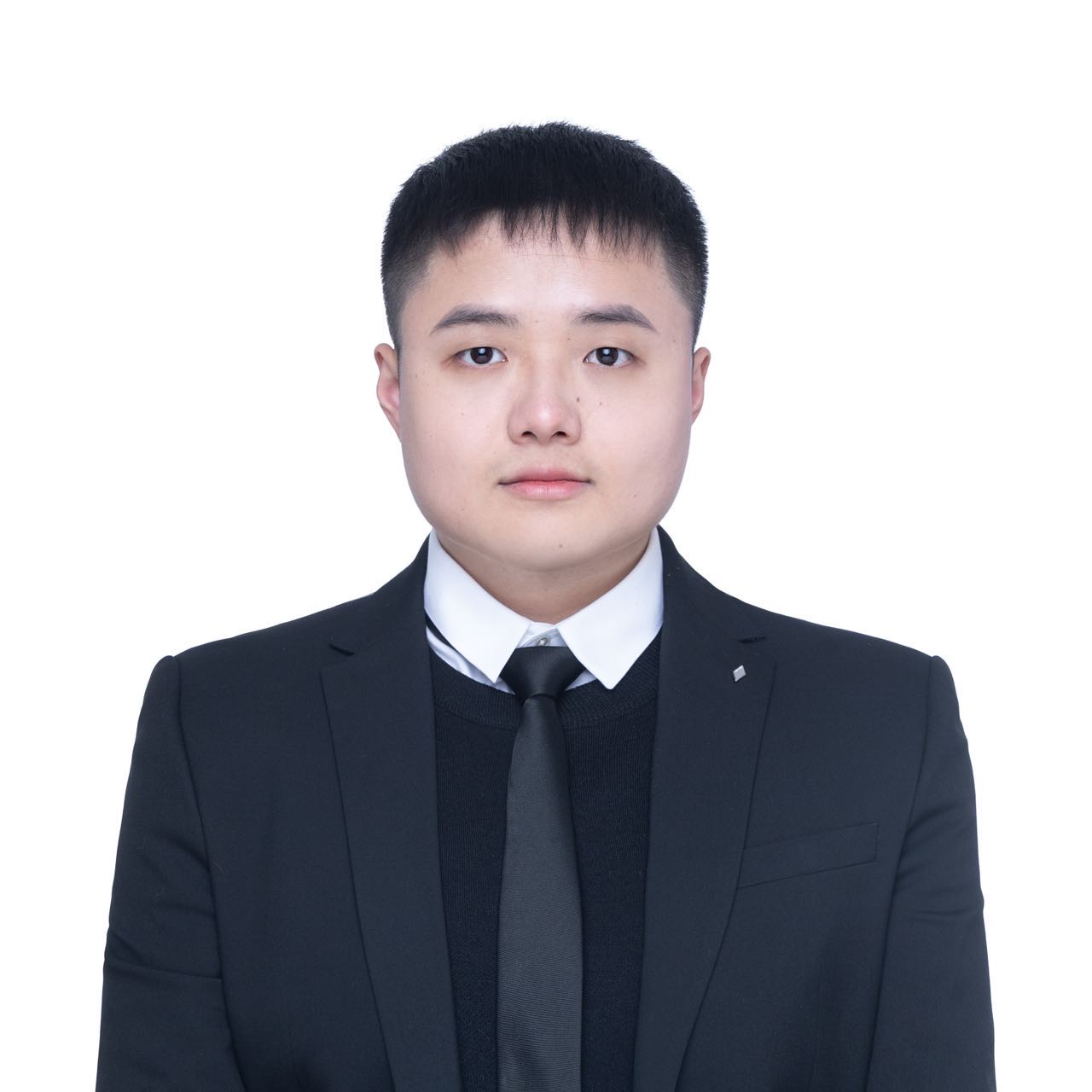}}]
{Kanglin Ning} received a B.S. degree from the Dalian University of Technology, Dalian, China, in 2016. and received the M.S. degree from the Department of Computer Science and Technology, the High-tech Institute of Xi'an. He is currently working toward a Ph.D. degree from the School of Computer Science, Harbin Institute of Technology (HIT), Harbin, China. His research interests include depth estimation, 3D object detection, vision occupancy, Multimodal Fusion.
\end{IEEEbiography}

\vspace{-1.5cm}
\begin{IEEEbiography}[{\includegraphics[width=1in,height=1.25in,clip,keepaspectratio]{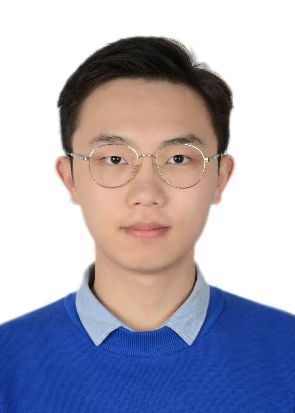}}]
{Wenrui Li} is currently an Associate Professor with the School of Computer Science, Harbin Institute of Technology. He received the B.S. degree from the School of Information and Software Engineering, University of Electronic Science and Technology of China (UESTC), Chengdu, China, in 2021, and the Ph.D. degree from the School of Computer Science, Harbin Institute of Technology (HIT), Harbin, China, in 2025. His research interests include multimedia search, low-level processing, and spiking neural networks. He was supported by the National Natural Science Foundation of China (NSFC) Youth Student Basic Research Program (Doctoral Student) in 2024. He has authored or co-authored more than 40 technical articles in refereed international journals and conferences.
\end{IEEEbiography}

\vspace{-1.5cm}
\begin{IEEEbiography}[{\includegraphics[width=1in,height=1.25in,clip,keepaspectratio]{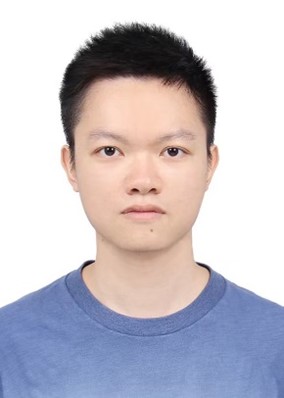}}]
{Houde Quan} received B.S. and M.S. degrees from Harbin Institute of Technology(HIT), China, in 2021 and 2023, respectively. He is currently working toward a Ph.D. degree from the School of Computer Science, Harbin Institute of Technology (HIT), Harbin, China. His research interests include computer vision and 3D scene understanding.
\end{IEEEbiography}

\vspace{-1.5cm}
\begin{IEEEbiography}[{\includegraphics[width=1in,height=1.25in,clip,keepaspectratio]{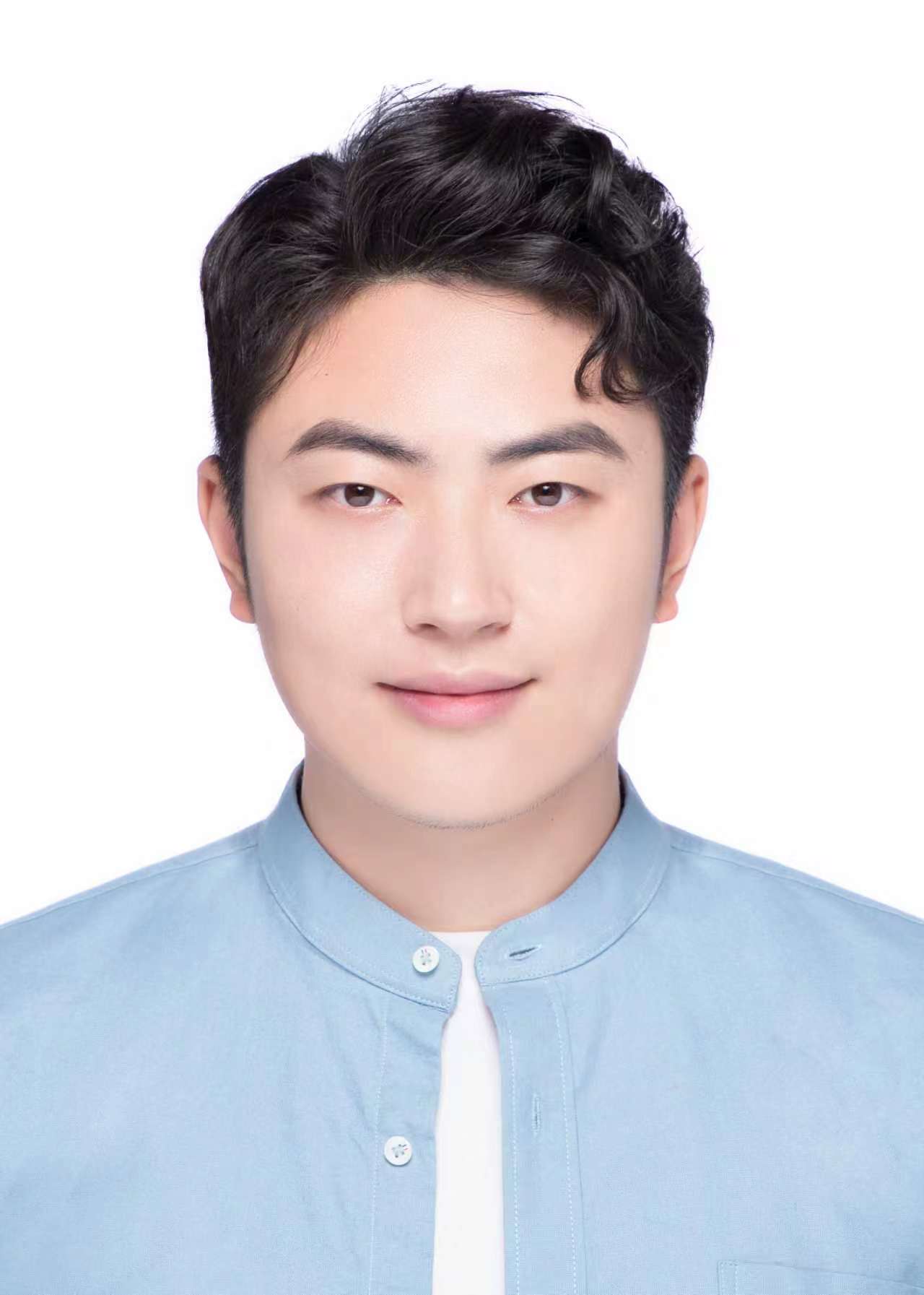}}]
{Qifan Li} received the B.S. degree in College of Computer Science and Technology from Huaqiao University, China, in 2020, and the M.S. degree in School of Computer Science from Fudan University, China, in 2023. He is currently pursuing the Ph.D. degree at the Faculty of Computing, Harbin Institute of Technology (HIT), Harbin, China. His research interests lie in low-level computer vision, including image and video restoration, image and video enhancement, and image and video super-resolution.
\end{IEEEbiography}

\vspace{-1.5cm}
\begin{IEEEbiography}[{\includegraphics[width=1in,height=1.25in,clip,keepaspectratio]{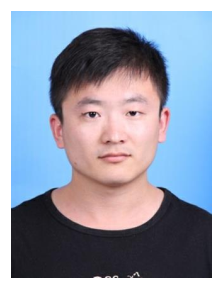}}]
{Xingtao Wang} received his B.S. degree from the Harbin Institute of Technology (HIT), Harbin, China, in 2016, and received the Ph.D. degree in computer science from HIT, Harbin, China, in 2022. From 2020 to 2022, he was with Peng Cheng Laboratory. He is currently a postdoc with the School of Computer Science and Technology, HIT. His research interests include point cloud denoising, mesh denoising, and deep learning.
\end{IEEEbiography}

\vspace{-1.5cm}
\begin{IEEEbiography}[{\includegraphics[width=1in,height=1.25in,clip,keepaspectratio]{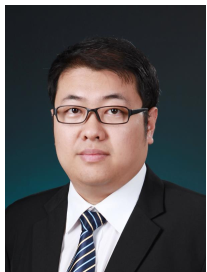}}]
{Xiaopeng Fan} (S'07-M'09-SM'17) received the B.S. and M.S. degrees from the Harbin Institute of Technology (HIT), Harbin, China, in 2001 and 2003, respectively, and the Ph.D. degree from The Hong Kong University of Science and Technology, Hong
Kong, in 2009. He joined HIT in 2009, where he is currently a Professor. From 2003 to 2005, he was with Intel Corporation, China, as a Software Engineer. From 2011 to 2012, he was with Microsoft Research Asia as a Visiting Researcher. From 2015 to 2016, he was with the Hong Kong University of Science and Technology as a Research Assistant Professor. He has authored one book and more than 100 articles in refereed journals and conference proceedings. His current research interests include video coding and transmission, image processing, and computer vision. He served as a Program Chair for PCM2017, Chair for IEEE SGC2015, and Co-Chair for MCSN2015. He was an Associate Editor of IEEE 1857 Standard in 2012. He received Outstanding Contributions to the Development of IEEE Standard 1857 by IEEE in 2013.
\end{IEEEbiography}

\end{document}